\documentclass[10pt,journal,compsoc]{IEEEtran}

\ifCLASSINFOpdf
\else
\usepackage[dvips]{graphicx}
\fi
\usepackage{url}

\usepackage[colorlinks,linkcolor=blue,anchorcolor=blue,
citecolor=blue]{hyperref}
\usepackage{graphicx}
\usepackage{amsmath,amssymb}
\usepackage{booktabs,tabularx,multirow,threeparttable}
\usepackage{float}
\usepackage{subfigure}
\usepackage{amssymb}
\usepackage{subfigure}
\ifCLASSOPTIONcompsoc
  \usepackage[nocompress]{cite}
\else
  \usepackage{cite}
\fi

\begin{document}
	\title{On the Arbitrary-Oriented Object Detection: Classification based Approaches Revisited
	}
	%~\IEEEmembership{Member,~IEEE,}
	\author{Xue~Yang, Junchi~Yan~\IEEEmembership{Senior Member,~IEEE}
		
% 		\thanks{Manuscript received April 19, 2005; revised August 26, 2015.}
		
		\thanks{Xue~Yang, Junchi~Yan (correspondence author) are with Department of Computer Science and Engineering, Shanghai Jiao Tong University, and also with the MoE Key Lab of Artificial Intelligence, AI Institute, Shanghai Jiao Tong University.}
		\thanks{E-mail: \{yangxue-2019-sjtu,yanjunchi\}@sjtu.edu.cn}
		}

	\markboth{Journal of \LaTeX\ Class Files,~Vol.~14, No.~8, August~2015}%
	{Shell \MakeLowercase{\textit{et al.}}: Bare Demo of IEEEtran.cls for IEEE Journals}

\IEEEtitleabstractindextext{
	\begin{abstract}
		Arbitrary-oriented object detection has been a building block for rotation sensitive tasks. We first show that the boundary problem suffered in existing dominant regression-based rotation detectors, is caused by angular periodicity or corner ordering, according to the parameterization protocol. We also show that the root cause is that the ideal predictions can be out of the defined range. Accordingly, we transform the angular prediction task from a regression problem to a classification one. For the resulting circularly distributed angle classification problem, we first devise a Circular Smooth Label technique to handle the periodicity of angle and increase the error tolerance to adjacent angles. To reduce the excessive model parameters by Circular Smooth Label, we further design a Densely Coded Labels, which greatly reduces the length of the encoding. Finally, we further develop an object heading detection module, which can be useful when the exact heading orientation information is needed e.g. for ship and plane heading detection. We release our OHD-SJTU dataset and OHDet detector for heading detection. Extensive experimental results on three large-scale public datasets for aerial images i.e. DOTA, HRSC2016, OHD-SJTU, and face dataset FDDB, as well as scene text dataset ICDAR2015 and MLT, show the effectiveness of our approach.
	\end{abstract}
	
	}
	\maketitle
	
	\section{Introduction}
	\IEEEPARstart{O}{bject} detection has been a standing task in computer vision. Recently, rotation detection has played an emerging and vital role in processing and understanding visual information from aerial images \cite{yang2021r3det, ding2018learning, yang2019scrdet, azimi2018towards, yang2018automatic}, scene text \cite{zhou2017east, liu2018fots, jiang2017r2cnn, ma2018arbitrary, liao2018rotation, liao2018textboxes++} and face \cite{shi2018real, huang2007high, rowley1998rotation}. The rotation detector can provide accurate orientation and scale information, which will be helpful in applications such as object change detection in aerial images and recognition of sequential characters for multi-oriented scene texts.
	
	\begin{figure*}[!tb]
	\centering
	\includegraphics[width=0.86\linewidth]{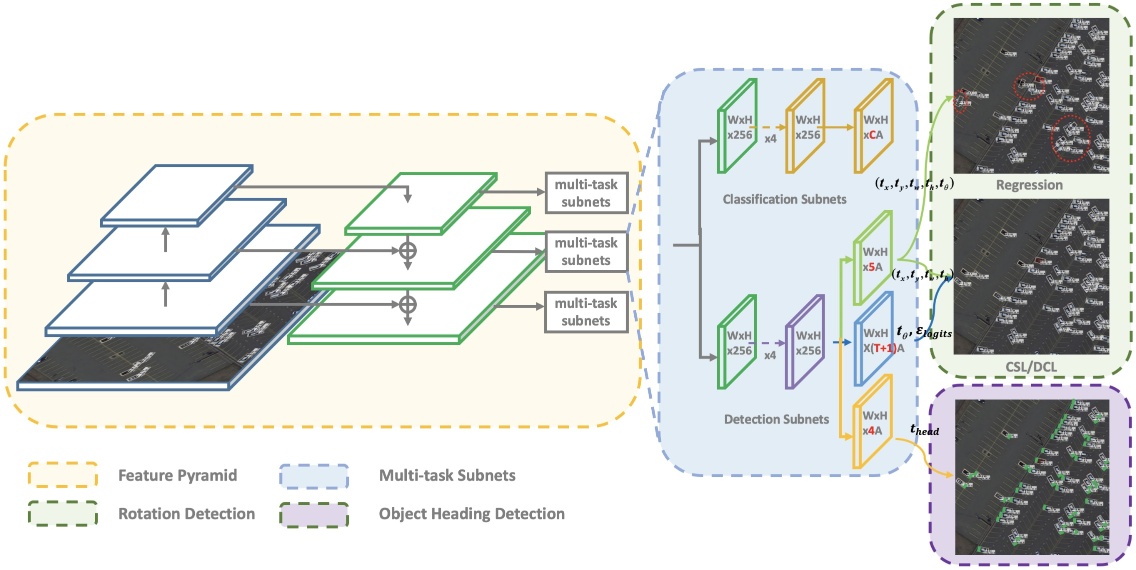}
	\caption{Architecture of the proposed detector (RetinaNet~\cite{lin2017focal} as an embodiment). `W' and `H' refer to the width and height respectively. `C' and `T' in red represents the number of object and encoding length of the angle, respectively. Object heading detection refers to further finding the head of the object based on the orientation information obtained by the rotation detection. Note that the CSL/DCL module on the right refers to the classification based prediction function either based on a Circular Smooth Label or the size reduced version Densely Coded Label.}
	\label{fig:pipeline}
	\vspace{-10pt}
\end{figure*}

Recently, a line of advanced rotation detectors evolved from classic detection algorithms \cite{girshick2015fast,ren2015faster,lin2017feature,lin2017focal,dai2016r} have been developed. Among these methods, detectors based on region regression take the dominance, and the representation of multi-oriented object is achieved by rotated bounding box or quadrangles. Although these rotation detectors have achieved promising results, there are still some fundamental problems. Specifically, we note both the five-parameter regression \cite{yang2019scrdet,ma2018arbitrary,ding2018learning} and the eight-parameter regression \cite{liao2018textboxes++, liu2019omnidirectional, qian2021learning, xu2020gliding} methods suffer the issue of discontinuous boundaries, as often caused by angular periodicity or corner ordering, depending on the choice of parameterization protocol. However, the root reasons are not limited to the particular representation of the bounding box. In this paper, we argue that the root cause of boundary problems based on regression methods is that the ideal predictions are beyond the defined range. Thus, the model's loss value suddenly increases at the boundary situation so that the model cannot obtain the prediction result in the simplest and most direct way, and additional complicated treatment is often needed. Therefore, these detectors often have difficulty in boundary conditions. For detection using rotated bounding boxes, the accuracy of angle prediction is critical. A slight angle deviation can lead to notable Intersection-over-Union (IoU) drop, resulting in inaccurate detection, especially for large aspect ratios. 
	
There have been some works addressing the boundary problem. For example, IoU-smooth L1 loss \cite{yang2019scrdet} introduces the IoU factor, and modular rotation loss \cite{qian2021learning} increases the boundary constraint to eliminate the sudden increase in boundary loss and reduce the difficulty of model learning. However, these regression-based detection methods still have not solved the root cause as mentioned above. 

In this paper, we aim to devise a more fundamental rotation detection baseline to solve the boundary problem. Specifically, we consider object angle prediction as a classification problem to better limit the prediction results, and then we design a Circular Smooth Label (CSL) to address angle periodicity and to increase the tolerance between adjacent angles. We show that the rotation accuracy error due to the conversion from continuous prediction to discrete bins can be negligible by a fine-granularity. We also introduce four window functions in CSL and explore the effect of different window radius sizes on detection performance. We further design two Densely Coded Labels (DCL), which greatly reduce the length of the encoding while ensuring the angle prediction accuracy does not sacrifice. In order to eliminate the theoretical prediction errors caused by angle dispersion, We also propose an angle fine-tuning mechanism.
Finally, we implement object heading detection on the basis of rotation detection, namely OHDet, and release a new dataset OHD-SJTU to the community. Through experiments and visual analysis, we show that CSL-based and DCL-based rotation detection algorithms are indeed better baseline choices than the angle regression-based protocol on different detectors and datasets. Note the regression-based, CSL-based and DCL-based protocols mentioned in the rest of the paper are named according to the prediction form of the angle. As a byproduct, we show how the object head can be effectively identified after detecting the rotating bounding box.

The preliminary content of this paper has partially appeared in ECCV 2020~\cite{yang2020arbitrary} and CVPR 2021~\cite{yang2021dense}\footnote{To obtain a more thorough analysis and comprehensive results, the conference versions~\cite{yang2020arbitrary, yang2021dense} have been significantly extended and improved in this journal version, especially in the following aspects: 
i) We explore the relationship between the angle discrete representation granularity denoted by $\omega$ and the detection performance. It shows that discrete granularity $\omega$ can be approximated as a CSL technique with a rectangular window function, which has a certain tolerance in the divided angle interval. The difference is that CSL smooths between adjacent angle intervals. See Table \ref{table:ablation_omega} in Section~\ref{sec:ablation_study}; ii) We use a specific calculation example to explain why the code length has such a large impact on the amount of detection model parameters and calculations, see Section \ref{sec:dcl}; 
iii) As for the angle prediction of the regression branch, we use two forms as the baseline to be compared, include direct regression and indirect regression, see Section \ref{sec:loss}; 
iv) We verify our approach on additional more challenging datasets, including FDDB, and DOTA-v1.5/v2.0, see Table \ref{table:other_dataset_comparison} and Table \ref{tab:dota1to2}. Among them DOTA-v1.5/v2.0 contain more data and tiny object (less than 10 pixels) than DOTA-v1.0;
v) We propose an angle fine-tuning mechanism to eliminate the theoretical prediction errors caused by angle dispersion which has been a common issue in whatever CSL and DCL, see Section \ref{sec:angle_fine_tune}; 
vi) As a common function for downstream applications, we develop a classification-based object heading detector in Section~\ref{sec:ohdet}. To verify its usefulness, we annotate and release a new dataset for this purpose and perform detection evaluation for both rotation and heading with a considerable amount, and more stringent evaluation indicators are used, as detailed in Section~\ref{sec:oh-sjtu}. To our best knowledge, this is the first public benchmark for multiple-category heading detection, especially at a considerable scale. Finally, we also release the full version of the source code.}. 
The overall contributions of this extended journal version can be summarized as:
\begin{itemize}
    \item We characterize the boundary problems encountered in different regression-based rotation detection methods~\cite{azimi2018towards, yang2018automatic, yang2021r3det, ding2018learning} and show the root cause is that the ideal predictions are beyond the defined range. %, which to our best knowledge, has not been identified and discussed before.
    
    \item To effectively dismiss the boundary problem, we design a novel classification based rotation detection paradigm, in contrast to the dominant regression based methods in existing works. The incurred accuracy error is negligible thanks to the devised fine-grained angle discretization (less than 1 degree) which to our best knowledge, has not been developed yet in the literature, and a coarse classification (around 10-degree) model is dated back to the 1990s for face detection~\cite{rowley1998rotation}.
    
    \item We develop the Circular Smooth Label (CSL) technique as an independent module. It can be readily reused in existing regression based methods by replacing the regression component with classification, to enhance angular prediction in face of boundary conditions and objects with large aspect ratio. We further design a Densely Coded Label (DCL) to solve the problem of excessive model parameters induced by CSL. DCL can greatly reduce the length of the encoding while maintaining a high angle prediction accuracy. We also propose an angle fine-tuning mechanism to eliminate the theoretical prediction errors caused by angle dispersion which has been a common issue in both CSL and DCL.
    
    \item On the basis of rotation detection, we further develop an object heading detector, namely OHDet, to identify the heading of object. In addition, we annotate and release a dataset called OHD-SJTU\footnote{\url{https://yangxue0827.github.io/OHD-SJTU.html}.}, which can be used for both rotation detection and object heading detection tasks.
		
	\item Extensive experimental results on HRSC2016 and DOTA-v1.0 show the state-of-the-art performance of our detector, and the efficacy of the CSL and DCL technique as independent components have been verified across different detectors. The source code \cite{yang2021alpharotate} is publicly available\footnote{\url{https://github.com/yangxue0827/RotationDetection}}. 
\end{itemize}

The paper is organized as follows. Section \ref{sec:related} introduces the related work. Section \ref{sec:method} presents the main approach in this paper and the experiments are conducted in Section \ref{sec:experiments}. Section \ref{sec:conclusion} concludes this paper.

\section{Related Work}\label{sec:related}
In this section, we discuss the related work, including classic region object detection and rotated object detection. Specifically, we discuss some relevant techniques regarding classification based orientation estimation, as well as the recent works on object heading detection.

\subsection{Horizontal Region Object Detection} 
Classic object detection aims to detect general objects with horizontal bounding boxes, and many high performance general-purpose object detectors have been proposed. R-CNN \cite{girshick2014rich} pioneers a method based on CNN detection. Subsequently, region-based models such as Fast R-CNN \cite{girshick2015fast}, Faster R-CNN \cite{ren2015faster}, and R-FCN \cite{dai2016r} are proposed, which improve the detection speed while reducing computational storage. FPN \cite{lin2017feature} focuses on the scale variance of objects in images and propose feature pyramid network to handle objects at different scales. SSD \cite{liu2016ssd}, YOLO \cite{redmon2016you} and RetinaNet \cite{lin2017focal} are representative single-stage methods, and their single-stage structure leads to higher detection speeds. Compared to anchor-based protocols, many anchor-free have become extremely popular in recent years. CornerNet \cite{law2018cornernet}, CenterNet \cite{duan2019centernet} and ExtremeNet \cite{zhou2019bottom} attempt to predict some keypoints of objects such as corners or extreme points, which are then grouped into bounding boxes. However, horizontal detector does not provide accurate orientation and scale information, which poses problem in real applications such as object change detection in aerial images and recognition of sequential characters for multi-oriented scene texts.
	
\subsection{Arbitrary-oriented Object Detection}
Aerial images and scene text are the main application scenarios of the rotation detector. Recent advances in multi-oriented object detection are mainly driven by adaption of classical object detection methods using rotated bounding boxes or quadrangles to represent multi-oriented objects. Due to the complexity of the remote sensing image scene and the large number of small, cluttered and rotated objects, multi-stage rotation detectors are still dominant for their robustness. Among them, ICN \cite{azimi2018towards}, ROI-Transformer \cite{ding2018learning}, SCRDet \cite{yang2019scrdet}, R$^3$Det \cite{yang2021r3det} are state-of-the-art detectors. Gliding Vertex \cite{xu2020gliding} and RSDet \cite{qian2021learning} achieve more accurate object detection through quadrilateral regression prediction. For scene text detection, RRPN \cite{ma2018arbitrary} employs rotated RPN to generate rotated proposals and further performs rotated bounding box regression. TextBoxes++ \cite{liao2018textboxes++} adopts vertex regression on SSD. RRD \cite{liao2018rotation}  further improves TextBoxes++ by decoupling classification and bounding box regression on rotation-invariant and rotation sensitive features, respectively. In fact, these mainstream regression-based methods often suffer the boundary problems due to the predictions beyond the defined range.

The idea of segmentation is an effective way of solving boundary problem. For example, segmentation-based protocols \cite{tian2019learning, liao2020real} are popular in the area of scene text detection. However, these methods are not practical for aerial images, which often contain a large number of densely arranged small objects in multiple categories. In contrast, instance segmentation is more suitable, such as Mask R-CNN \cite{he2017mask}, SOLO \cite{wang2020solo}, and CondInst \cite{tian2020conditional}, but there are also many limitations.

First, Considering that instance segmentation requires a lot of labeling workload, a more straightforward solution is to convert the rotated boxes into binary masks \cite{wang2019mask}. However, such conversion will introduce many background areas, which will reduce the classification accuracy of pixels and affect the accuracy of the final prediction box. Besides, for the top-down methods (e.g. Mask RCNN), dense scenes will limit the detection of horizontal boxes because of the excessive suppression of dense horizontal overlapping bounding boxes due to non-maximum suppression (NMS), thereby affecting subsequent segmentation. Last but not least, the bottom-up methods, such as SOLO and CondInst, assign different instances to different channels, so they are not suitable for aerial images, which often show large scale scenes with a large number of dense and small objects. Take the parking lot scene in DOTA-v1.5 \cite{xia2018dota} dataset as an example, a sub-image with a size of $450\times600$ will contain up to 2,000 vehicles, which are often less than 10 pixels in size and are densely arranged.

The above reasons may help explain why angle-based rotation detection algorithms still dominate in aerial imagery which is an important application area. Therefore, we design a new rotation detection baseline, which basically eliminates the boundary problem by transforming angle prediction from a regression problem to a classification problem.

\subsection{Classification for Orientation Information} Early works have been developed for multi-view face detection with arbitrary rotation-in-plane (RIP) angles, by obtaining orientation information via classification. Specifically, the divide-and-conquer technique is adopted in \cite{huang2007high}, which uses several small neural networks to deal with a small range of face appearance variations individually. In \cite{rowley1998rotation}, a router network is firstly used to estimate each face candidate’s RIP angle. PCN \cite{shi2018real} progressively calibrates the RIP orientation of each face candidate and shrinks the RIP range by half in early stages. Finally, PCN makes the accurate final decision for each face candidate to determine whether it is a face and predict the precise RIP angle. In other research areas, \cite{kim2019instance} adopts ordinal regression for effective future motion classification. \cite{yang2018position} obtains the orientation information of the ship by classifying the four sides. 
The above methods all obtain the approximate orientation range through classification, but cannot be directly applied to scenarios that require precise orientation information such as aerial images and scene text. They also do not suffer the boundary problem due to the PoA and EoE, because their prediction granularity is very rough and the aspect ratio of object is small.

\subsection{Object Heading Detection} 
DLR 3K \cite{liu2015fast} is an aerial image dataset that can be used for car head detection. There are 20 images in total, including 3,418 cars and 54 trucks. For car head detection, the authors first use a sliding window strategy with binary classification to detect cars, and then perform a rough estimation of the head through classification, with 16 classes ($22.5^\circ$ rotation difference between adjacent sample groups, respectively). This method needs to train multiple detectors and cannot perform high-precision angle prediction. DRBox \cite{liu2017learning} and DRBox-v2 \cite{an2019drbox} define the rotation bounding box according to the head of the object, and the angle predicted by regression can be used to determine the direction of the object head. However, the bounding box of this definition protocol has a relatively large angular range, at $[0^\circ,360^\circ)$, which is challenging. EAGLE \cite{azimi2021eagle} is a large-scale dataset for vehicle detection in aerial imagery, which has still not been released so far. This paper releases a new dataset  called OHD-SJTU with labeled heading information, and it covers more categories of objects.

\begin{figure}[!tb]
	\centering
	\subfigure[]{
		\begin{minipage}[t]{0.3\linewidth}
			\centering
			\includegraphics[width=1.\linewidth]{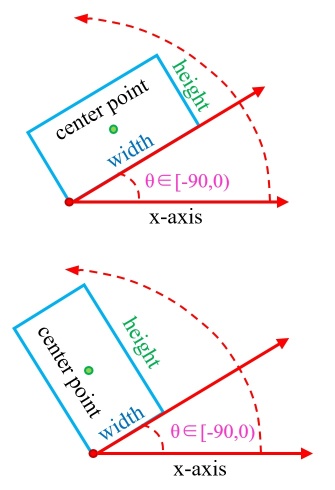}
			%\caption{fig1}
		\end{minipage}%
		\label{fig:bbox_90}
	}
	\subfigure[]{
		\begin{minipage}[t]{0.3\linewidth}
			\centering
		\includegraphics[width=1.\linewidth]{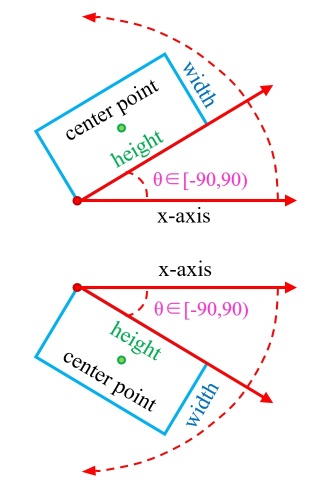}
			%\caption{fig2}
		\end{minipage}%
		\label{fig:bbox_180}
	}
	\subfigure[]{
		\begin{minipage}[t]{0.3\linewidth}
			\centering
			\includegraphics[width=1.\linewidth]{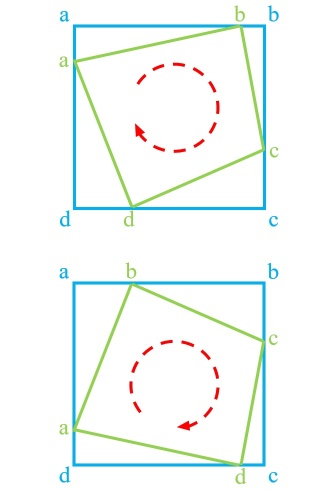}
			%\caption{fig2}
		\end{minipage}
		\label{fig:quad}
	}
	\centering
	\caption{Popular definitions of bounding boxes in existing literature. (a) Five-parameter method with $90^\circ$ angular range \cite{yang2019scrdet, yang2021r3det}. (b) Five-parameter method with $180^\circ$ angular range \cite{ma2018arbitrary, ding2018learning}. (c) Ordered quadrilateral representation \cite{liao2018textboxes++, liu2019omnidirectional, qian2021learning, xu2020gliding}.}
	\label{fig:bbox_defination}
	\vspace{-10pt}
\end{figure}

In summary, this work is dedicated to proposing a general detection method that can be used for multi-class high-precision rotation detection and object heading detection.

\section{Proposed Approach}\label{sec:method}
Figure \ref{fig:pipeline} gives an overview of our method. The embodiment is a single-stage rotation detector based on the RetinaNet~\cite{lin2017focal}. The figure shows a multi-tasking pipeline, including classification branch, rotation detection branch and object heading detection branch. Among them, rotation detection branch contains regression based prediction and CSL-based prediction, to facilitate the comparison of the performance of the two methods. It can be seen from the figure that CSL-based protocol is more accurate for learning the orientation and scale information of the object. The DCL-based protocol maintains consistent performance with CSL. 

Note that the method proposed in this paper is applicable to most regression-based protocols by replacing the regression module with our classification one.

\subsection{Regression-based Rotation Detection Method}
Parametric regression is currently a popular method for rotation object detection, mainly including five-parameter regression-based protocols \cite{yang2018automatic, yang2019scrdet, yang2021r3det, ma2018arbitrary, ding2018learning, jiang2017r2cnn} and eight-parameter regression-based protocols \cite{liao2018textboxes++, liu2019omnidirectional, qian2021learning, xu2020gliding}. The commonly used five-parameter regression-based protocols realize arbitrary-oriented bounding box detection by adding an additional angle parameter $\theta$. Figure \ref{fig:bbox_90} shows one of the rectangular definition $(x,y,w,h,\theta)$ with $90^\circ$ angular range \cite{yang2018automatic, yang2019scrdet, yang2021r3det, ma2018arbitrary}, $\theta$ denotes the acute angle to the x-axis, and for the other side we refer it as $w$. It should be distinguished from another definition $(x,y,h,w,\theta)$ illustrated in Figure \ref{fig:bbox_180}, with $180^\circ$ angular range \cite{ding2018learning, ma2018arbitrary}, whose $\theta$ is determined by the long side ($h$) of the rectangle and x-axis. The eight-parameter regression-based detectors directly regress the four corners $(x_1,y_1,x_2,y_2,x_3,y_3,x_4,y_4)$ of the object, so the prediction is a quadrilateral. The key step to the quadrilateral regression is to sort the four corner points in advance, which  avoids large loss even if the prediction is correct, as shown in Figure \ref{fig:quad}.

\begin{figure}[!tb]
	\centering
	\subfigure[The $90^\circ$-regression-based protocol]{
		\begin{minipage}[t]{0.98\linewidth}
			\centering
			\includegraphics[width=1.\linewidth]{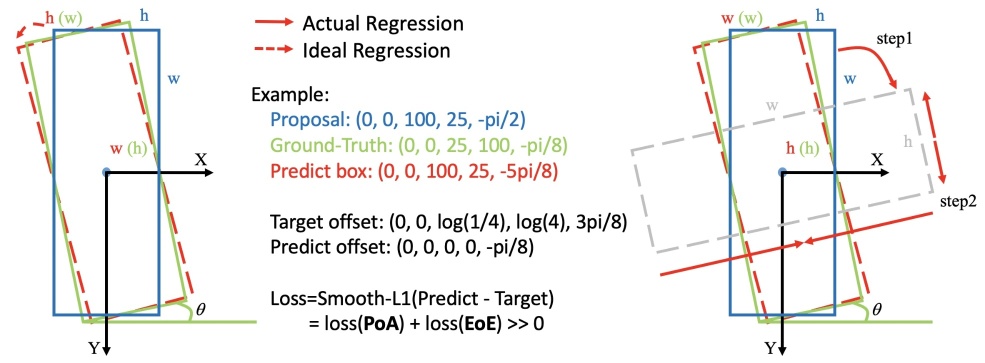}
			%\caption{fig1}
		\end{minipage}%
		\label{fig:problem_90}
	}\\
	\subfigure[The $180^\circ$-regression-based protocol]{
		\begin{minipage}[t]{0.95\linewidth}
			\centering
			\includegraphics[width=1.\linewidth]{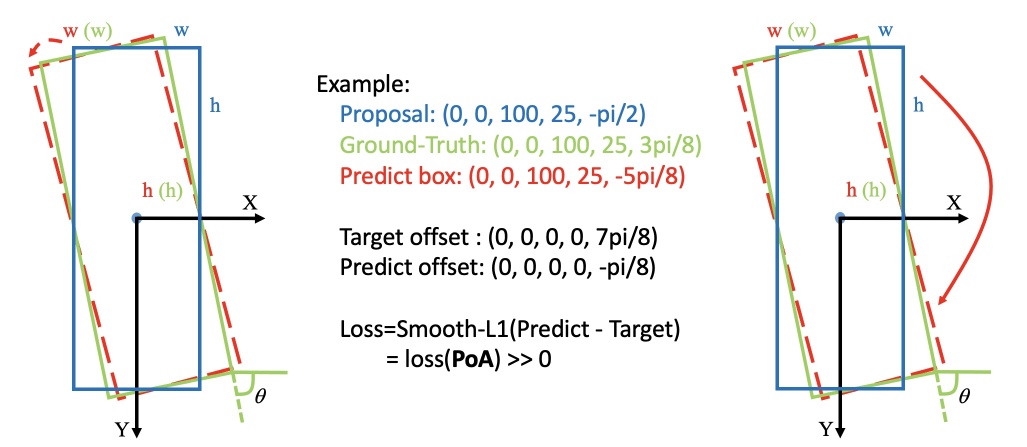}
			%\caption{fig2}
		\end{minipage}%
		\label{fig:problem_180}
	}\\
	\subfigure[Point-regression-based protocol]{
		\begin{minipage}[t]{0.85\linewidth}
			\centering
			\includegraphics[width=1.\linewidth]{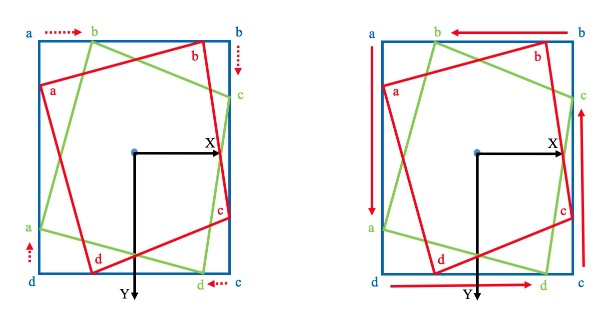}
			%\caption{fig2}
		\end{minipage}
		\label{fig:problem_quad}
	}\\
	\centering
	\caption{Illustration for the boundary problem which persistently exists over three popular categories of regression based protocols. The red solid arrow indicates the actual regression process, and the red dotted shows the ideal regression process.}
	\label{fig:problem}
\end{figure}

\subsection{Boundary Problem of Regression Method}\label{sec:bp}
Although the parametric regression-based rotation detection method has achieved competitive performance in visual detection, these methods essentially suffer the discontinuous boundaries problem \cite{yang2019scrdet,qian2021learning}. On the surface, boundary discontinuity problems are often caused by angular periodicity under the five-parameter protocol, or the corner ordering in the eight-parameter setting. While there exists a more fundamental root cause behind the representation choice of the bounding box.

\begin{figure*}[!tb]
	\centering
	\subfigure[RetinaNet-H\cite{yang2021r3det}]{
		\begin{minipage}[t]{0.18\linewidth}
			\centering
			\includegraphics[width=0.93\linewidth]{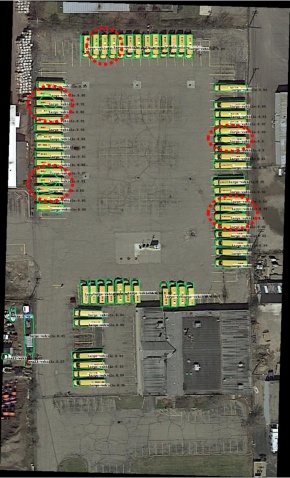}
			%\caption{fig1}
		\end{minipage}%
		\label{fig:retinanet-h}
	}
	\subfigure[FPN-H \cite{lin2017feature}]{
		\begin{minipage}[t]{0.18\linewidth}
			\centering
			\includegraphics[width=0.93\linewidth]{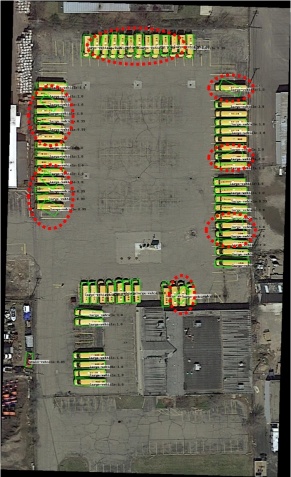}
			%\caption{fig2}
		\end{minipage}
		\label{fig:fpn-h}
	}
	\subfigure[R$^3$Det\cite{yang2021r3det}]{
		\begin{minipage}[t]{0.18\linewidth}
			\centering
			\includegraphics[width=0.93\linewidth]{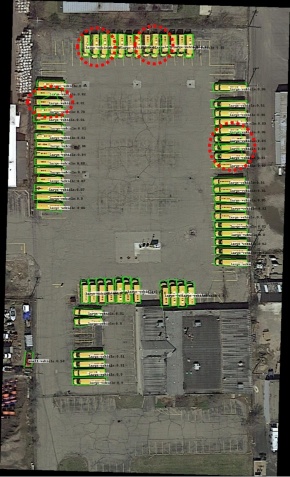}
			%\caption{fig1}
		\end{minipage}%
		\label{fig:r3det}
	}
	\subfigure[IoU-Smooth L1\cite{yang2019scrdet}]{
		\begin{minipage}[t]{0.18\linewidth}
			\centering
			\includegraphics[width=0.93\linewidth]{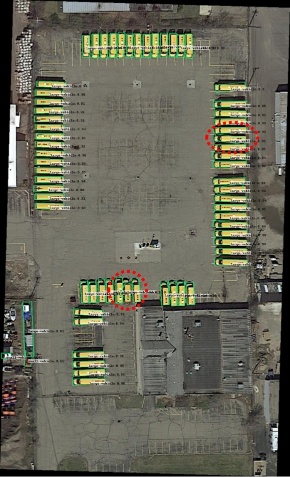}
			%\caption{fig2}
		\end{minipage}%
		\label{fig:iou-smooth-l1}
	}
	\subfigure[$180^\circ$-DCL]{
		\begin{minipage}[t]{0.18\linewidth}
			\centering
			\includegraphics[width=0.93\linewidth]{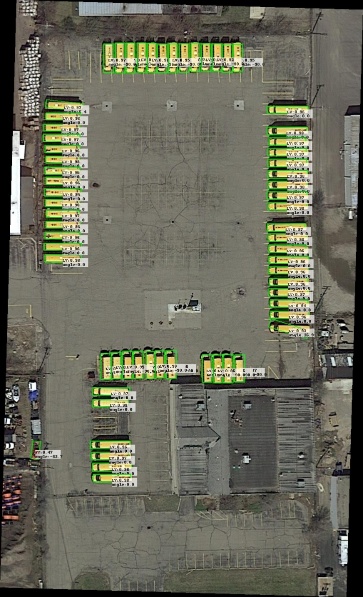}
			%\caption{fig2}
		\end{minipage}%
		\label{fig:gcl-180}
	}\\
	\vspace{-8pt}
	\subfigure[$180^\circ$-CSL-Pulse]{
		\begin{minipage}[t]{0.18\linewidth}
			\centering
			\includegraphics[width=0.93\linewidth]{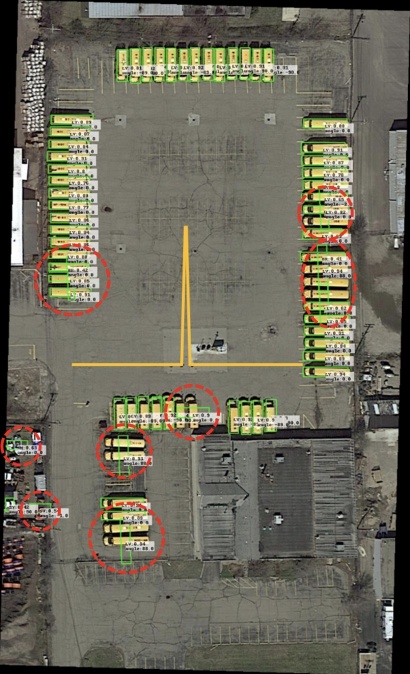}
			%\caption{fig2}
		\end{minipage}
		\label{fig:csl-pulse}
	}
	\subfigure[$180^\circ$-CSL-Rect.]{
		\begin{minipage}[t]{0.18\linewidth}
			\centering
			\includegraphics[width=0.93\linewidth]{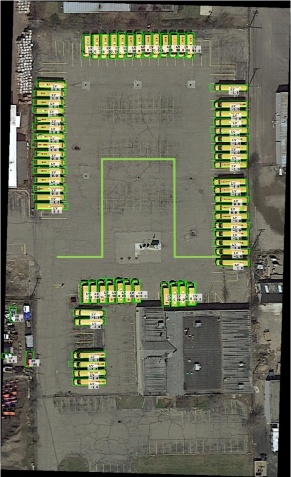}
			%\caption{fig2}
		\end{minipage}
		\label{fig:csl-rect}
	}
	\subfigure[$180^\circ$-CSL-Triangle]{
		\begin{minipage}[t]{0.18\linewidth}
			\centering
			\includegraphics[width=0.93\linewidth]{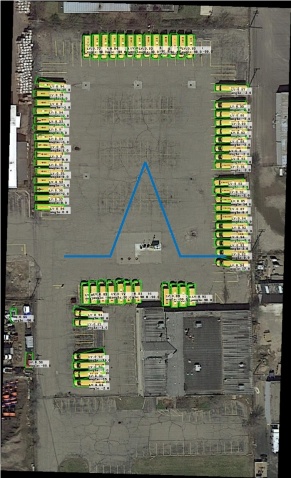}
			%\caption{fig2}
		\end{minipage}
		\label{fig:csl-tria}
	}
	\subfigure[$180^\circ$-CSL-Gaussian]{
		\begin{minipage}[t]{0.18\linewidth}
			\centering
			\includegraphics[width=0.93\linewidth]{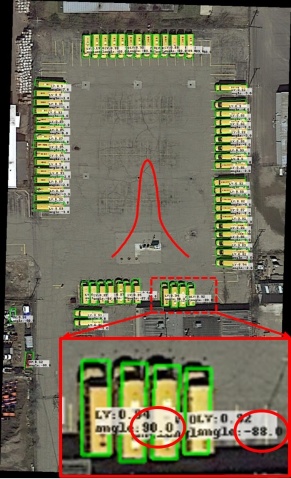}
			%\caption{fig2}
		\end{minipage}
		\label{fig:csl-gauss-180}
	}
	\subfigure[$90^\circ$-CSL-Gaussian]{
		\begin{minipage}[t]{0.18\linewidth}
			\centering
			\includegraphics[width=0.93\linewidth]{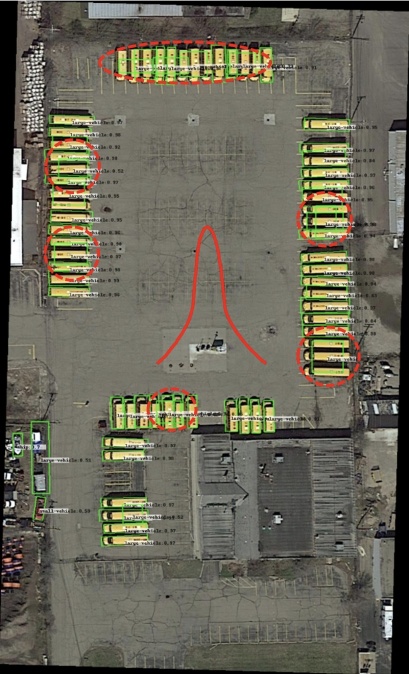}
			%\caption{fig2}
		\end{minipage}
		\label{fig:csl-gauss-90}
	}
	\centering
	\caption{Comparison of four regression-based rotation detection methods and angle classification-based protocols in the boundary case.  `H' and `R' represent the horizontal and rotating anchors. Red dotted circles indicate some bad cases.  Figures \ref{fig:csl-pulse}-\ref{fig:csl-gauss-180} show that the Gaussian window function performs best, while the pulse function performs worst because it has not learned any orientation and scale information. According to Figure \ref{fig:csl-gauss-180} and Figure \ref{fig:csl-gauss-90}, the $180^\circ$-CSL-based protocol obviously has better boundary prediction due to the EoE problem still exists in the $90^\circ$-CSL-based protocol. In general, CSL-based and DCL-based protocols have no boundary problem, as shown in Figure \ref{fig:csl-gauss-180} and \ref{fig:gcl-180}.}
	\label{fig:win_func_vis}
\end{figure*}

The boundary discontinuity can cause the loss value suddenly increase at the boundary situation. Thus methods have to resort to particular and often complex tricks to mitigate this issue. Therefore, these detection methods are often inaccurate in boundary conditions. We describe the boundary problem in three typical categories of regression-based protocols according to their different representation forms (the first two refer to the five-parameter methods):
	
\begin{itemize}
	\item \textbf{The $90^\circ$-regression-based protocol, as sketched in Figure \ref{fig:problem_90}.} It shows that an ideal form of regression (the blue box rotates counterclockwise to the red box), but the loss of this situation is very large due to the periodicity of angular (PoA) and exchangeability of edges (EoE), see the example in Figure \ref{fig:problem_90} and Equation \ref{eq:regression1}, \ref{eq:regression2}, \ref{eq:multitask_loss} for detail. Therefore, the model has to be regressed in other complex forms (such as the blue box rotating clockwise to the gray box while scaling w and h), increasing the difficulty of regression. It should be noted that the prediction box and ground truth in the ideal regression way do have a high IoU value in visual perception, but the prediction box at this time has exceeded our defined range so that we cannot calculate the accurate IoU if no additional judgment processing is performed.
	
	\item \textbf{The $180^\circ$-regression-based protocol, as illustrated in Figure \ref{fig:problem_180}.} Similarly, this method also suffers the issue of sharp increase of loss caused by the PoA at the boundary. The model will eventually choose to rotate the proposal a large angle clockwise to get the final predicted bounding box.
	
	\item \textbf{Point-regression-based protocol, as shown in Figure \ref{fig:problem_quad}.} Through further analysis, the boundary discontinuity problem still exists in the eight-parameter regression method due to the advance ordering of corner points. Consider the situation of an eight-parameter regression in the boundary case, the ideal regression process should be $\{({\color{blue}{a}}\rightarrow {\color{green}{b}}),({\color{blue}{b}}\rightarrow {\color{green}{c}}),({\color{blue}{c}}\rightarrow {\color{green}{d}}),({\color{blue}{d}}\rightarrow {\color{green}{a}})\}$, but the actual regression process from the blue reference box to the green ground truth box is $\{({\color{blue}{a}}\rightarrow {\color{green}{a}}),({\color{blue}{b}}\rightarrow {\color{green}{b}}),({\color{blue}{c}}\rightarrow {\color{green}{c}}),({\color{blue}{d}}\rightarrow {\color{green}{d}})\}$. In fact, this situation also belongs to PoA. By contrast, the actual and ideal regression of the blue to red bounding boxes is consistent.
\end{itemize}

Some approaches have been proposed to solve these problems based on the above analysis. For example, IoU-smooth L1 \cite{yang2019scrdet} loss introduces the IoU factor, and modular rotation loss \cite{qian2021learning} increases the boundary constraint to eliminate the sudden increase in boundary loss and reduce the difficulty of model learning. However, these methods are still regression-based detection methods, and no solution is given from the root cause.
	
In this paper, we will start from a new perspective and replace regression with classification to achieve better and more robust rotation detectors. We reproduce some classic rotation detectors based on regression and compare them visually under boundary conditions, as shown in Figure \ref{fig:retinanet-h} to Figure \ref{fig:iou-smooth-l1}. In contrast, CSL-based and DCL-based protocols have no boundary problem, as shown in Figure \ref{fig:csl-gauss-180} and \ref{fig:gcl-180}.

\subsection{Vanilla Angular Classification}
The main cause of boundary problems based on regression methods is that the ideal predictions are beyond the defined range. Therefore, we consider the prediction of object angle as a classification task to restrict the prediction range. One simple solution is to use the object angle as its category label, and the number of categories is related to the angle range. Figure \ref{fig:one-hot_label} shows the label setting for a vanilla classification problem (one-hot label encoding). The conversion from regression to classification can cause certain accuracy error. Taking the five-parameter method with $180^\circ$ angle range as an example: $\omega$ (default $\omega=1^\circ$) degree per interval refers to a category for labeling. It calculates the maximum accuracy error $Max (error)$ and the expected accuracy error $E(error)$:
\begin{equation}
%\small
	\begin{aligned}
	Max(error) =& \frac{\omega}{2} \\
	E(error) = &\int_{a}^{b} \frac x{b-a} dx = \int_{0}^{\frac{\omega}{2}} \frac x{\frac{\omega}{2}-0} dx = \frac\omega{4}
	\label{eq:error}
	\end{aligned}
\end{equation}
where $\omega=AR/C_{\theta}$ indicates the angle discretization granularity. $AR$ and $C_{\theta}$ represents angle range (the default value is 180) and the number of angle categories, respectively.  

Based on the above equations, one can see the error is slight for a rotation detector with small enough angle discrete granularity $\omega$. For example, when two rectangles with a $1:9$ aspect ratio differ by $0.25^\circ$ and $0.5^\circ$ (default expected and maximum accuracy error), the Intersection over Union (IoU) between them only decreases by 0.02 and 0.05. 

The discrete equation and prediction equation of the angle are as follows:
\begin{equation}
%\small
	\begin{aligned}
	\textbf{Encode:} & \ \text{One-Hot}( -\text{Round}((\theta_{gt}-90)/\omega)) \\
	 \textbf{Decode:} & \  90 - \omega \left(0.5 + \text{Argmax}(\text{Sigmoid}(logits))\right)
	\label{eq:discrete}
	\end{aligned}
\end{equation}
where $\theta_{gt}$ presents the angle decimal label.

Applying vanilla classification methods to prediction of angles is appeared earlier in the field of face detection \cite{huang2007high, rowley1998rotation, shi2018real}. As these works only need approximate orientation range, e.g.10-degree ($\omega=10$) in \cite{rowley1998rotation}. We find that vanilla angular classification methods are difficult to deal with objects with multiple categories and large aspect ratio, e.g. DOTA dataset. In contrast, the small aspect ratio and single category characteristics in face detection make it unnecessary for high-precision angle prediction, so high-precision angle classification still has not been solved.

\subsection{Circular Smooth Label for Angular Classification}\label{sec:csl}
In our analysis, there are two reasons why vanilla classification methods cannot obtain high-precision angle prediction for rotation detection: 

\textbf{Reason i)} The EoE problem still exists when the bounding box uses the $90^\circ$-based protocol, as shown in Figure \ref{fig:csl-gauss-90}. Moreover, $90^\circ$-based protocol has two different border cases (vertical and horizontal), while $180^\circ$-based protocol has only vertical border cases. %Therefore, we use $180^\circ$-regression-based protocol in this paper.

\begin{figure}[!tb]
	\centering
	\subfigure[One-hot label.]{
		\begin{minipage}[t]{0.44\linewidth}
			\centering
			\includegraphics[width=1\linewidth]{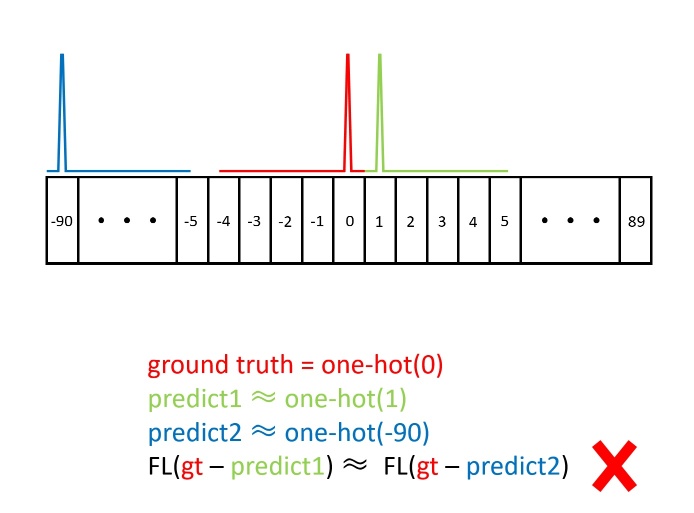}
		\end{minipage}%
		\label{fig:one-hot_label}
	}
	\subfigure[Circle smooth label.]{
		\begin{minipage}[t]{0.44\linewidth}
			\centering
			\includegraphics[width=1\linewidth]{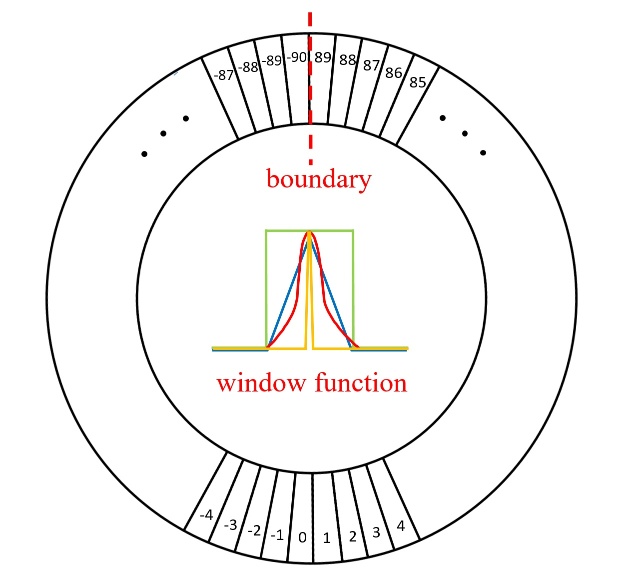}
		\end{minipage}%
		\label{fig:circle_label}
	}\\
	\centering
	\caption{Two kinds of labels for angular classification. FL means using the focal loss function \cite{lin2017focal}.}
	\label{fig:label}
\end{figure}

\textbf{Reason ii)}  Note vanilla classification loss is agnostic to the angle distance between the predicted label and ground truth label, thus it is inappropriate for the nature of the angle prediction problem. As shown in Figure \ref{fig:one-hot_label}, when the ground truth is $0^\circ$ and the prediction results of the classifier are $1^\circ$ and $-90^\circ$ respectively, their prediction losses are the same, but the prediction results close to ground truth should be allowed from a detection perspective.

Therefore, Circular Smooth Label (CSL) technique is designed to obtain more robust angular prediction through classification without suffering boundary conditions, including EoE and PoA. It should be noted that CSL can only solve the PoA, and the EoE problem can be solved by the $180^\circ$ angular definition method. It can be clearly seen from Figure \ref{fig:circle_label} that CSL involves a circular label encoding with periodicity, and the assigned label value is smooth with a certain tolerance. The expression of CSL is as follows:
\begin{equation}
%\small
    \begin{aligned}
    CSL(x) = 
    \left\{ \begin{array}{rcl}
    g(x), & \theta-r<x<\theta+r \\ 0, & otherwise
    \end{array}\right.
    \label{eq:csl}
    \end{aligned}
\end{equation}
where $g(x)$ is a window function. $r$ is the radius of the window function. $\theta$ represents the angle of the current bounding box. An ideal window function $g(x)$ is required to hold the following properties:
\begin{itemize}
	\item \textbf{Periodicity}: $g(x)=g(x+kT), k \in N$. $T=180/\omega$ represents the number of bins into which the angle is divided, and the default value is 180.
	\item \textbf{Symmetry}: $0 \leq g(\theta+\varepsilon)=g(\theta-\varepsilon) \leq 1,  |\varepsilon|<r$. $\theta$ is the center of symmetry.
	\item \textbf{Maximum}: $g(\theta)=1$.
	\item \textbf{Monotonic}: $0 \leq g(\theta \pm \varepsilon)\leq g(\theta \pm \varsigma) \leq 1,  |\varsigma|<|\varepsilon|<r$. The function presents a monotonous non-increasing trend from the center point to both sides
\end{itemize}

Figure \ref{fig:circle_label} shows four efficient window functions that meet the above four properties: pulse functions, rectangular functions, triangle functions, and Gaussian functions. Note that the label value is continuous at the boundary and there is no arbitrary accuracy error due to the periodicity of CSL. In addition, one-hot label (vanilla classification) is equivalent to CSL when the window function is a pulse function or the radius of the window function is very small. Equation \ref{eq:encode_decode_csl} describes the angle prediction process in CSL:
\begin{equation}
%\small
	\begin{aligned}
	 \textbf{Encode:} & \ \text{CSL}( -\text{Round}((\theta_{gt}-90)/\omega)) \\
	 \textbf{Decode:} & \ 90 - \omega (\text{Argmax}(\text{Sigmoid}(logits)) + 0.5)
	\label{eq:encode_decode_csl}
	\end{aligned}
\end{equation}

\subsection{Densely Coded Label for Angular Classification}\label{sec:dcl}
The CSL-based detectors adopt the so-called Sparsely Coded Label (SCL) encoding technique.
Although CSL has many good properties for angle prediction, the The design of CSL will cause the prediction layer to introduce too much parameter and calculation, resulting in inefficiency of the detector. Specifically, the One-Hot and CSL described above are sparse angle encoding methods, which often leads to excessively long angle encoding length:
\begin{equation}
	\begin{aligned}
	L_{one-hot}=  L_{csl}= AR / \omega
	\label{eq:len_csl}
	\end{aligned}
\end{equation}

\begin{figure}[!tb]
	\centering
	\includegraphics[width=0.9\linewidth]{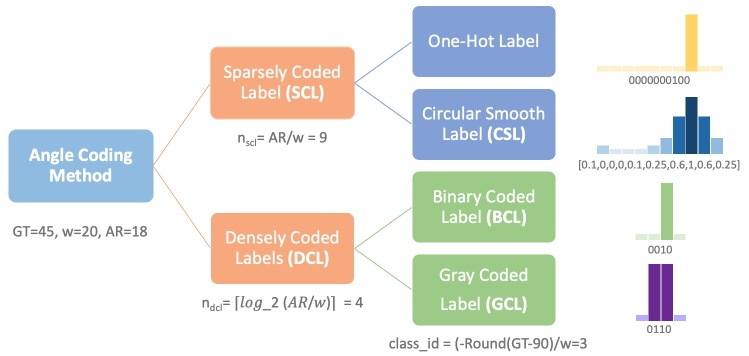}
	\caption{The relationship between the various angle encoding methods.}
	\label{fig:coding_relation}
\end{figure}

Binary Coded Label (BCL) \cite{heath1972origins} and Gray Coded Label (GCL) \cite{frank1953pulse} are two Densely Coded Label (DCL) methods commonly used in the field of electronic communication. Their advantage is that they can represent a larger range of values with less coding length. Thus, they can effectively solve the problem of excessively long coding length in CSL and One-Hot based methods. BCL processes the angle by binarization to obtain a string of codes represented by multiple `0' and `1'. In the encoding of a group of numbers, if any two adjacent codes differ only by one binary number, then this kind of encoding is called Gray Code. In addition, because only one digit is different between the maximum number and the minimum number, it is also called Cyclic Code. The encoding forms between adjacent angles are not much different, which makes GCL also have a certain classification tolerance. The cycle characteristics of GCL are also consistent with circular design idea of CSL. Table \ref{table:coding} compares the coding results of BCL and GCL and Figure \ref{fig:coding_relation} shows the relationship between various angle encoding methods.

\begin{table}[tb!]
	\centering
	\caption{The three-digit binary code and gray code corresponding to the decimal number.}
	\resizebox{0.48\textwidth}{!}{
		\begin{tabular}{ccccccccccc}
			\toprule
			Decimal Number & 0 & 1 & 2 & 3 & 4 & 5 & 6 & 7\\
			\midrule
			
			Binary Coded Label & 000 & 001 & 010 & 011 & 100 & 101 & 110 & 111 \\
			Gray Coded Label & 000 & 001 & 011 & 010 & 110 & 111 & 101 & 100 \\

			\bottomrule
			
	\end{tabular}}
	\label{table:coding}
\end{table}

The code length of DCL is:
\begin{equation}
    \begin{aligned}
    L_{dcl} = \left\lceil \log_{2}({AR / \omega}) \right\rceil
    \label{eq:len_gcl}
    \end{aligned}
\end{equation}

We will use a specific calculation example to explain why the code length has such a large impact on the amount of detection model parameters and calculations. Take RetinaNet as an example, the number of prediction layer channels can be calculated as follows:
\begin{equation}
    \begin{aligned}
    C_{o} = A \times L
    \end{aligned}
\end{equation}
where $A$ represents the number of anchors, and $A=scale\_num \times ratio\_num \times angle\_num$. In this paper $scale\_num=3, ratio\_num=7, angle\_num=1$.

For all prediction layers \{P3, P4, P5, P6, P7\}, the total Flops and Params are calculated as follows:
\begin{equation}
\begin{aligned}
Flops =& \sum_{i=3}^{7}Flops_{i}=\sum_{i=3}^{7}2C_{i}^{i}K_{i}^{2}H_{i}W_{i}C_{o}^{i}\\
Params=&C_{i}K^{2}C_{o}
\end{aligned}
\end{equation}
where $K_{i}$, $H_{i}$ and $W_{i}$ denote the convolution kernel size of the $i$-th level prediction layer, and the height and width of the input feature map for the $i$-th level prediction layer. $C_{i}$ and $C_{o}$ represent the number of input and output channels of the $i$-th level prediction layer. It should be noted that the parameters of different levels of prediction layers are shared.

\begin{table}[tb!]
\caption{Comparison of GFlops and Param over rotation detectors, under the same setting and hyperparameters. The baseline is RetinaNet.}
\centering
\resizebox{0.48\textwidth}{!}{
	\begin{tabular}{cccrcrcc}
		\toprule
		Method & $\omega$ & GFlops & $\Delta$GFlops & Params & $\Delta$Params & Training & Inference \\
		\midrule
		
		Reg. & - & 139.35 & - & 36.97 M & - & - \\
		CSL & 1 & 254.96 & +82.96\% & 45.63 M & +23.42\% & $\sim$1/3x & $\sim$1/2x \\
		GCL & 1 & 143.87 & \textbf{+3.24\%} & 37.31 M & \textbf{+0.92\%} & \textbf{$\sim$1x} & \textbf{$\sim$1x} \\
		\bottomrule
	\end{tabular}}
	\label{table:flops_param}
\end{table}

According to the default settings of our paper, the input image size is $800$, $C_i=256$, $C_o=AL=21L$, $K$=3. Then, taking $AR=180$, $w=1$ as an example, the code length required by CSL and One-Hot are $L_{onehot}=L_{csl}=180$, while the code length of DCL is only $L_{dcl}=8$. The total Flops and Params of all prediction layers are $1,291,175,424L$ and $48,384L$. The huge base make the prediction layer occupy the main calculations and parameters of the model. Therefore, the shortening of the code length is necessary and important. Finally, we have counted the total parameters and calculations of three different models, as shown in Table \ref{table:flops_param}. From the perspective of GFlops and Params, detectors based on CSL have increased by about 82.96\% and 45.63\%, respectively. In contrast, DCL-based protocol only increases by 3.24\% and 0.92\%. The training and testing time of RetinaNet-DCL is about 3 times and 2 times faster than RetinaNet-CSL, respectively.

% BCL processes the angle by binarization to obtain a string of codes represented by multiple `0' and `1'. Although the coding length is greatly reduced, there may be huge changes in the coding results between adjacent values, that is, there is no classification tolerance mentioned in the CSL. For example, the three-bit binary coding results of the values `3' and `4' are `011' and `100', respectively. It can be seen that all three positions have changed, resulting in a very large difference in the loss value of the two angle predictions. GCL can solve this problem. In the encoding of a group of numbers, if any two adjacent codes differ only by one binary number, then this kind of encoding is called Gray Code. Due to only one digit is different between the maximum number and the minimum number, it is also called Cyclic Code. The coding results of `3' and `4' in the GCL method are `010' and `110'. Table \ref{table:coding} compares the coding results of BCL and GCL. The shortcomings of GCL are also obvious. Although the encoding forms between adjacent angles are not much different, which makes GCL also have a certain classification tolerance, the encoding differences of angles with large differences are not very significant, such as `1 (001)' and `6 (101)'. In summary, these two methods are agnostic or partially agnostic to the angle distance.

In the DCL-based method, only the number of categories is a power of 2 to ensure that each coding corresponds to a valid angle. For example, if the 180 degree range is divided into $2^8=256$ categories, then the range of each division interval is $\omega=180/256=0.703125^\circ$. According to the $Max(error) = \omega / 2$ and $E(error)=\omega / 4$, the maximum and expected accuracy error are only $0.3515625^\circ$ and $0.17578125^\circ$, whose influence on final detecton accuracy can be negligible. However, the above condition is not necessary. We find that even with some redundant invalid codes, there is no significant drop in final performance. Equation \ref{eq:encode_decode_dcl} specifies the encoding and decoding process of DCL (take BCL as an example):
\begin{equation}
%\small
	\begin{aligned}
	 \textbf{Encode:} & \ \text{Bin}( -\text{Round}((\theta_{gt}-90)/\omega)) \\
	 \textbf{Decode:} & \ 90 - \omega \text{Int}(\text{Round}(\text{Sigmoid}(logits)))
	\label{eq:encode_decode_dcl}
	\end{aligned}
\end{equation}

\begin{figure}[!tb]
	\centering
	\includegraphics[width=0.98\linewidth]{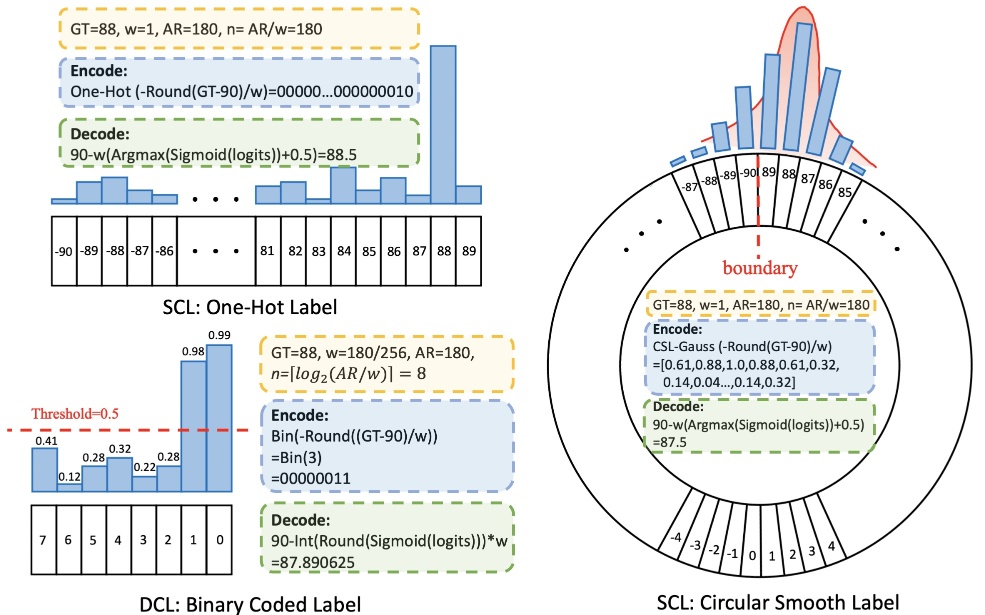}
	\caption{Examples of encoding and decoding process of One-Hot, CSL-Gaussian and BCL for angle prediction.}
	\label{fig:angle_label}
\end{figure}

Figure \ref{fig:angle_label} gives the examples of encoding and decoding process of One-Hot, CSL-Gaussian and BCL for angle prediction.

\subsection{Angle Fine-Tuning}\label{sec:angle_fine_tune}
A larger angle discrete granularity $\omega$ can alleviate the problem of too many parameters in the prediction layer to a certain extent, and reduce the dependence on the classification ability of the model angle. However, the theoretical angle error in Eq. \ref{eq:error} cannot be ignored at this time. To solve this problem, we predict a smaller angle $\varepsilon$ to make up for the error of accuracy caused by angle encoding.
\begin{equation}
    \begin{aligned}
    \theta_{disc} = & \text{Decode}(\text{Encode}(\theta_{gt})) \\
    \varepsilon_{gt} = & \theta_{gt} - \theta_{disc}, \quad \varepsilon_{gt} \in \left[-\frac{\omega}{2}, \frac{\omega}{2}\right]
    \end{aligned}
    \label{eq:theta_error}
\end{equation}

Taking RetinaNet in Figure \ref{fig:pipeline} as an example, we will add an extra one-dimensional output at the prediction layer, denoted as $\varepsilon_{logits}$, and then fine-tune the angle by:
\begin{equation}
    \begin{aligned}
    \varepsilon_{pred}=&\left(\text{Sigmoid}(\varepsilon_{logits})-0.5\right)*\omega \\
    \theta_{pred}^{'} =& \min(\max(\theta_{pred} + \varepsilon_{pred}, -90^\circ), 90^\circ)
    \end{aligned}
\end{equation}
where $\theta_{pred}$ and $\theta_{pred}^{'}$ respectively represent the predicted angle before and after fine-tuning the angle. The $\min$ and $\max$ operations are to avoid PoA.

\subsection{Object Heading Detector}\label{sec:ohdet}
\begin{figure}[!tb]
	\centering
	\subfigure[Heading label is 0.]{
		\begin{minipage}[t]{0.46\linewidth}
			\centering
			\includegraphics[width=0.95\linewidth]{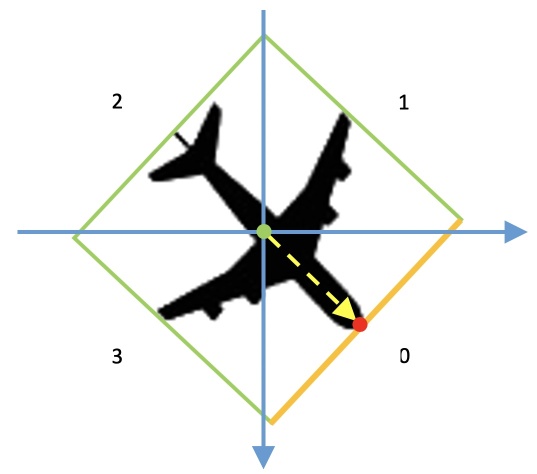}
		\end{minipage}%
		\label{fig:head_label_0}
	}
	\subfigure[Heading label is 1.]{
		\begin{minipage}[t]{0.46\linewidth}
			\centering
			\includegraphics[width=0.95\linewidth]{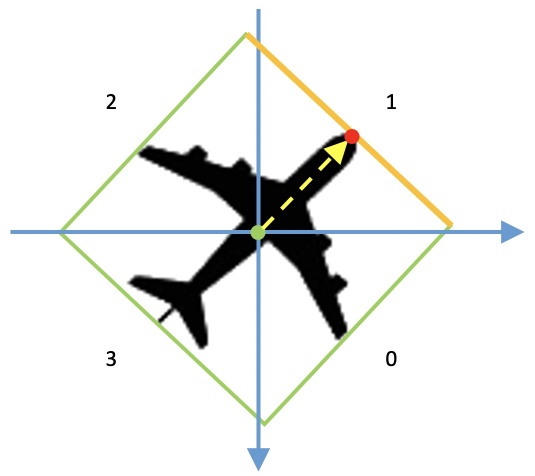}
		\end{minipage}%
		\label{fig:head_label_1}
	}\\ \vspace{-5pt}
	\subfigure[Heading label is 0.]{
		\begin{minipage}[t]{0.46\linewidth}
			\centering
			\includegraphics[width=0.95\linewidth]{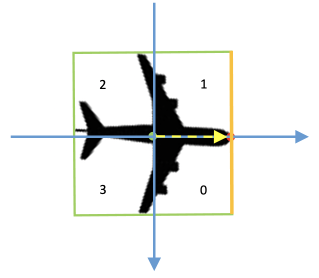}
		\end{minipage}%
		\label{fig:head_label_0_h}
	}
	\subfigure[Heading label is 1.]{
		\begin{minipage}[t]{0.46\linewidth}
			\centering
			\includegraphics[width=0.95\linewidth]{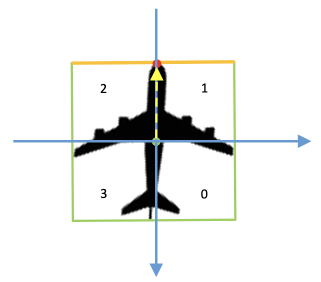}
		\end{minipage}%
		\label{fig:head_label_1_v}
	}
	\centering
	\caption{Definition of object heading label. The green bounding box shows the  rotation detection. The red dot denotes the head of the object. The green dot indicates the center of the rotating bounding box. The yellow dotted line shows the object's head orientation, and the coordinate quadrant pointed by the yellow dotted line is the heading label.}
	\label{fig:head_label}
\end{figure}

Compared with rotation detection, object heading detection is a more fine-grained detection task, which aims to determine the head of the object. Although the rotation detection retains the orientation information of the object, it still cannot determine the accurate head of the object based on the angle of the rotating bounding box alone. By a carefully study, we find that the head of the object must be located in the four sides of the rotating bounding box. Inspired by this discovery, we only need to perform an additional simple four-category to predict the head. Figure \ref{fig:head_label} shows how the object heading label is defined.

One essential prerequisite for realizing object heading detection is an accurate rotation detector, which is expected to satisfy the following two characteristics:
\begin{itemize}
	\item In non-boundary situations, the detector can output high-precision rotating bounding boxes.
	\item In boundary situations, the detector is not sensitive to boundary problem.
\end{itemize}

Therefore, cascade multi-stage strategy and angle classification technique can be combined to solve the above problems. Based on the above analysis, we have adjusted the entire detector as follows:

\begin{itemize}
	\item The number of anchors plays a vital role in the performance of the detector, which can be calculated by $num\_scales \times num\_ratios \times num\_angles$. Therefore, we do not use any angle classification technique (e.g. CSL or DCL) in the first stage, so that an appropriate number of anchors can be set in the first stage to provide high-quality initial candidate boxes.
	\item After using regression prediction to obtain the refined anchor in the first stage, each feature point on the feature map only retains the refined anchor with the highest confidence. The filtering of the refined anchor makes the parameter A is always equal to 1 at each refined stage, which means that we can use angle classification technique.
	\item We use 90-degree angle definition method, as shown in Figure \ref{fig:bbox_90}, to reduce $AR$ to a minimum. At the same time, in order to solve the unprocessed boundary problem in the first stage and the EoE problem in the refinement stage, we use the IoU-Smooth L1 loss function \cite{yang2019scrdet} in each stage.
	\item Experimental results show that proper adjacent angle prediction fault tolerance can improve the performance of the detector. When the evaluation standard is not too strict (e.g. DOTA uses AP$_{50}$ as the evaluation metric), we can appropriately increase $w$ and adjust the radius $r$ of the window function to relieve the pressure on the prediction layer.

\end{itemize}

Combining the head prediction strategy and the above procedures, an efficient object heading detection method is fulfilled, which is called OHDet\footnote{\url{https://github.com/SJTU-Thinklab-Det/OHDet_Tensorflow}.}.

\subsection{Loss Function Design}\label{sec:loss}
Our multi-tasking pipeline contains regression-based prediction branch and angle classification-based prediction branch, to facilitate the performance comparison of the two methods on an equal footing. The model mainly outputs four items for location and size:
\begin{equation}
\begin{aligned}
t_{x}^{'}&=(x_{}^{'}-x_{a})/w_{a}, t_{y}^{'}=(y_{}^{'}-y_{a})/h_{a}, \\
t_{w}^{'}&=\log(w_{}^{'}/w_{a}), t_{h}^{'}=\log(h_{}^{'}/h_{a}) %t_{\theta}^{'}&=(\theta_{}^{'}-\theta_{a})\cdot\pi/180 \quad (only\;for\;reg.\;branch)
\label{eq:regression2}
\end{aligned}
\end{equation}
to match the four targets from the ground truth:
\begin{equation}
\begin{aligned}
t_{x}&=(x-x_{a})/w_{a}, t_{y}=(y-y_{a})/h_{a}, \\
t_{w}&=\log(w/w_{a}), t_{h}=\log(h/h_{a})
%t_{\theta}&=(\theta-\theta_{a})\cdot\pi/180 \quad (only\;for\;reg.\;branch),
\label{eq:regression1}
\end{aligned}
\end{equation}
where $x,y,w,h$ denote the box's center coordinates, width, height and angle, respectively. Variables $x, x_{a}, x^{'}$ are for the ground truth box, anchor box, and predicted box, respectively (likewise for $y,w,h$).

As for the angle prediction of the regression branch, we use two forms as the baseline to be compared:
\begin{itemize}
	\item Direct regression (\textbf{Reg.}). The model directly predicts the angle offset $t_{\theta}^{'}$:
	\begin{equation}
    \begin{aligned}
    	t_{\theta}=&(\theta-\theta_{a})\cdot\pi/180, \\ t_{\theta}^{'}=&(\theta_{}^{'}-\theta_{a})\cdot\pi/180
    \end{aligned}
    \end{equation}
	\item Indirect regression (\textbf{Reg.$^*$}). The model predicts two vectors ($t_{\sin\theta}^{'}$ and $t_{\cos\theta}^{'}$) to match the two targets from the ground truth ($t_{\sin\theta}$ and $t_{\cos\theta}$):
	\begin{equation}
    \begin{aligned}
        t_{\sin\theta} =& \sin{(\theta\cdot\pi/180)}, t_{\cos\theta} = \cos{(\theta\cdot\pi/180)}, \\ t_{\sin\theta}^{'}=&\sin{(\theta^{'}\cdot\pi/180)}, t_{\cos\theta}^{'}=\cos{(\theta^{'}\cdot\pi/180)}
    \end{aligned}
    \end{equation}
\end{itemize}
To ensure that $t_{\sin\theta}^{'2}+t_{\cos\theta}^{'2}=1$ is satisfied, we will perform the following normalization processing:
\begin{equation}
%\small
\begin{aligned}
t_{\sin\theta}^{'}=\frac{t_{\sin\theta}^{'}}{\sqrt{t_{\sin\theta}^{'2}+t_{\cos\theta}^{'2}}}, \ t_{\cos\theta}^{'}=\frac{t_{\cos\theta}^{'}}{\sqrt{t_{\sin\theta}^{'2}+t_{\cos\theta}^{'2}}}
\end{aligned}
\end{equation}

Indirect regression is a simpler way to avoid boundary problems. The multi-task loss is defined as follows:
\begin{equation}
%\small
\begin{aligned}
L = & \frac{\lambda_{1}}{N_{pos}}\sum_{n=1}^{N_{pos}}L_{reg}(t_{n}^{'},t_{n}) \\	
& + \frac{\lambda_{2}}{N_{pos}}\sum_{n=1}^{N_{pos}} L_{angle\_cls}(\theta_{n}^{'},\theta_{n}) \\
& + \frac{\lambda_{3}}{N_{pos}}\sum_{n=1}^{N_{pos}} L_{reg}(\varepsilon_{gt}, \varepsilon_{pred})\\ 
& + \frac{\lambda_{4}}{N_{pos}}\sum_{n=1}^{N_{pos}}L_{head}(h_{n}^{'}, h_{n}) + \frac{\lambda_{5}}{N}\sum_{n=1}^{N}L_{cls}(p_{n},l_{n}) 
\label{eq:multitask_loss}
\end{aligned}
\end{equation}
where $N$ and $N_{pos}$ indicate the total number of samples and positive samples, respectively. $t_{n}^{'}$ denotes the predicted vectors, $t_{n}$ is the targets vector of ground truth. $\theta_{n}$ and $\theta_{n}^{'}$ denote the label and prediction of angle, respectively. $\varepsilon_{gt}$ and $\varepsilon_{pred}$ represent the theoretical angle error and the predicted angle error. $h_{n}$ and $h_{n}^{'}$ represent the head of ground truth and prediction bounding box, respectively. $l_{n}$ represents the label of object, $p_{n}$ is the probability distribution of various classes calculated by sigmoid function. The hyper-parameter $\lambda_{k}$ ($k=1,2,..5$) control the trade-off and are set to $\{1,0.5,20,0.1,1\}$ by default. The classification loss $L_{cls}$, $L_{head}$ and $L_{angle\_cls}$ are focal loss \cite{lin2017focal} or sigmoid cross-entropy loss depends on the detector. The regression loss $L_{reg}$ is smooth L1 loss as used in \cite{girshick2015fast}.

\section{Experiments}\label{sec:experiments}
We use Tensorflow \cite{abadi2016tensorflow} to implement the proposed methods on a server with GeForce RTX 2080 Ti and 11G memory. The experiments in this article are initialized by ResNet50 \cite{he2016deep} by default unless otherwise specified. We perform experiments on both aerial benchmarks and scene text benchmarks to verify the generality of our techniques. Weight decay and momentum are set 0.0001 and 0.9, respectively. We employ MomentumOptimizer over 4 GPUs with a total of 4 images per minibatch (1 image per GPU). At each pyramid level, we use anchors at seven aspect ratios $\{1,1/2,2,1/4,4,1/6,6\}$, and the settings of the remaining anchor numbers are the same as the original RetinaNet and FPN.

\subsection{Benchmarks and Protocls}\label{sec:oh-sjtu}
\textbf{DOTA} \cite{xia2018dota} is a complex aerial image dataset for object detection, which contains objects exhibiting a wide variety of scales, orientations, and shapes. DOTA-v1.0 contains 2,806 aerial images and 15 common object categories from different sensors and platforms. The fully annotated DOTA-v1.0 benchmark contains 188,282 instances, each of which is labeled by an arbitrary quadrilateral. There are two detection tasks for DOTA: horizontal bounding boxes (HBB) and oriented bounding boxes (OBB). The training set, validation set, and test set account for 1/2, 1/6, 1/3 of the entire data set, respectively. 
In contrast, DOTA-v1.5 uses the same images as DOTA-v1.0, but extremely small instances (less than 10 pixels) are also annotated. Moreover, a new category, containing 402,089 instances in total is added in this version. While DOTA-v2.0 contains 18 common categories, 11,268 images and 1,793,658 instances. Compared to DOTA-v1.5, it includes the new categories. The 11,268 images in DOTA-v2.0 are split into training, validation, test-dev, and test-challenge sets.
We divide the images into $ 600 \times 600 $ subimages with an overlap of 150 pixels and scale it to $ 800 \times 800 $. With all these processes, we obtain about 27,000 patches. 
% The model is trained by 135k iterations in total, and the learning rate changes during the 81k and 108k iterations from 5e-4 to 5e-6. 
The short names for categories are defined as (abbreviation-full name): PL-Plane, BD-Baseball diamond, BR-Bridge, GTF-Ground field track, SV-Small vehicle, LV-Large vehicle, SH-Ship, TC-Tennis court, BC-Basketball court, ST-Storage tank, SBF-Soccer-ball field, RA-Roundabout, HA-Harbor, SP-Swimming pool, HC-Helicopter, CC-container crane, AP-airport and HP-helipad.

\textbf{ICDAR2015} \cite{karatzas2015icdar} is the Challenge 4 of ICDAR 2015 Robust Reading Competition, which is commonly used for oriented scene text detection and spotting. This dataset includes 1,000 training images and 500 testing images. In training, we first train our model using 9,000 images from ICDAR 2017 MLT training and validation datasets, then we use 1,000 training images to fine-tune our model.

\textbf{ICDAR 2017 MLT} \cite{nayef2017icdar2017} is a multi-lingual text dataset. It includes 7,200 training images, 1,800 validation images and 9,000 testing images. The dataset is composed of complete scene images in 9 languages, and text regions can be in arbitrary orientations, being more diverse and challenging.

\textbf{HRSC2016} \cite{liu2017high} contains images from two scenarios: ships on sea and ships close inshore. All images are collected from six harbors around the world. The training, validation and test set include 436, 181 and 444 images, respectively.

\textbf{FDDB} \cite{jain2010fddb} is a dataset designed for unconstrained face detection, in which faces have a wide variability of face scales, poses, and appearance. This dataset contains annotations for 5,171 faces in a set of 2,845 images taken from the faces in the Wild dataset~\cite{berg2005s}. In our paper, we manually use 70\% as the training set and the rest as the validation set.

\begin{table}[tb!]
    \centering
    \caption{Statistics on the categories and quantities of OHD-SJTU datasets. Abbreviations of the object categories are in brackets.}
    \label{tab:statistics}
    \resizebox{0.48\textwidth}{!}{
        \begin{tabular}{c|c|c|c|c|c|c}
        \toprule
         \multirow{2}{*}{Tag} & Plane & Ship & Small vehicle & Large vehicle & Harbor & Helicopter \\
        %\cline{2-7}
        & (PL) & (SH) & (SV) & (LV) & (HA) & (HC) \\
        \hline
        L train & 8,614 & 30,386 & 26,126 & 16,969 & 5,983 & 630 \\
        L val & 2,754 & 9,985 & 5,438 & 4,387 & 2,090 & 73 \\
        \hline
        S train & 559 & 2,318 & - & - & - & - \\
        S val & 223 & 1,025 & - & - & - & - \\
        \bottomrule
        \end{tabular}}
\end{table}

\begin{table}[tb!]
    \centering
    \caption{Comparison between different angle prediction methods on DOTA-v1.0 validation set. For CSL-based protocol, the label mode and windows function radius are set to Gaussian and 6, respectively. The baseline is RetinaNet. The angle range is $[-90^\circ,90^\circ)$.}
    \label{tab:angle_pred}
    \resizebox{0.48\textwidth}{!}{
        \begin{tabular}{c|c|cc|c}
        \toprule
       Angle Pred. & $\omega$ & AP$_{50}$ & AP$_{75}$ & AP$_{50:95}$ \\
        \hline
      Reg. ($\Delta \theta$) & - & 62.21 & 26.06 & 31.49 \\
         Reg.$^*$ ($\sin{\theta}$, $\cos{\theta}$) & - & 63.23 & 30.63 & 33.19 \\
          Cls. CSL & 1 & 64.40 & 32.58 & 35.04 \\
          Cls. BCL & 180/256 & 65.93 & \textbf{35.66} & \textbf{36.71}\\
         Cls. GCL & 180/256 & \textbf{66.13} & 33.65 & 36.34\\
        \bottomrule
        \end{tabular}}
\end{table}

\begin{table}[!tb]
	\centering
	\caption{Comparison of the four window functions (with radius set to 6) on the DOTA-v1.0 test set. 5-mAP refers to the mean average precision of the five categories with large aspect ratio. mAP means mean average precision of all 15 categories. The base method is RetinaNet-CSL. Note the EoE problem exists for the angle range $[-90^\circ,0^\circ)$.}
	\resizebox{0.49\textwidth}{!}{
		\begin{tabular}{c|c|c|c|c|c|c|c|c}
			\toprule
			Angle Range  &  Label Mode
			& BR & SV & LV & SH & HA & 5-mAP & mAP \\
			\hline
			\multirow{4}{*}{$[-90^\circ,0^\circ)$}  & Pulse & 9.80 & 28.04 & 11.42 & 18.43 & 23.35 & 18.21 & 39.52 \\
		& Rectangular & 37.62 & 54.28 & 48.97 & 62.59 & 50.26 & 50.74 & 58.86 \\
			& Triangle & 37.25 & 54.45 & 44.01 & 60.03 & 52.20 & 49.59 & 60.15 \\
		& Gaussian & \textbf{41.03} & \textbf{59.63} & \textbf{52.57} & \textbf{64.56} & \textbf{54.64} & \textbf{54.49} & \textbf{63.51} \\
			\cline{1-9}
			\multirow{4}{*}{$[-90^\circ,90^\circ)$}  & Pulse & 13.95 & 16.79 & 6.50 & 16.80 & 22.48 & 15.30 & 42.06 \\
			&  Rectangular & 36.14 & 60.80 & 50.01 & 65.75 & \textbf{53.17} & 53.17 & 61.98 \\
			&  Triangle & 32.69 & 47.25 & 44.39 & 54.11 & 41.90 & 44.07 & 57.94 \\
			&  Gaussian & \textbf{40.55} & \textbf{66.77} & \textbf{51.50} & \textbf{73.60} & 46.05 & \textbf{55.69} & \textbf{65.69} \\
			\bottomrule
	\end{tabular}}
	\label{table:ablation_window_func}
\end{table}

\begin{table}[tb!]
	\centering
	\caption{Comparison of detection mAP under different radius on the DOTA-v1.0 test set. The angle range, label mode and $\omega$ are set to $[-90^\circ,90^\circ)$, Gaussian and 1, respectively.}
	\resizebox{0.48\textwidth}{!}{
		\begin{tabular}{c|c|c|c|c|c}
			\toprule
			Method & r=0 & r=2 & r=4 & r=6 & r=8 \\
			\hline
			RetinaNet-CSL & 40.78 & 59.23 & 62.12 & \textbf{65.69} & 63.99 \\
			FPN-CSL & 48.08 & 70.18 & 70.09 & \textbf{70.92} & 69.75 \\
			\bottomrule
	\end{tabular}}
	\label{table:ablation_radius}
	\vspace{-10pt}
\end{table}

\begin{table}[tb!]
	\centering
	\caption{Comparison of detection results under different angle discrete granularity $\omega$ on the DOTA-v1.0 test set. The angle range is $[-90^\circ,90^\circ)$. For CSL-based protocol, the label mode and windows function radius are set to Gaussian and 1, respectively.}
	\resizebox{0.48\textwidth}{!}{
		\begin{tabular}{c|c|c|c|c|c}
			\toprule
			Granularity & $\omega$=30 & $\omega$=18 & $\omega$=10 & $\omega$=3 & $\omega$=1 \\
			\hline
			%Vanilla RetinaNet-H & 180 & Pulse & & & & & 42.06\\
			RetinaNet-CSL & 40.81 & 66.10 & \textbf{67.38}  & 64.81 & 58.92\\
			\hline
			Granularity & $\omega$=180/4 & $\omega$=180/32 & $\omega$=180/64 & $\omega$=180/128 & $\omega$=180/256 \\
			\hline
			RetinaNet-GCL & 62.38 & 65.59 & \textbf{67.02} & 65.14 & 64.97 \\
			\bottomrule
	\end{tabular}}
	\label{table:ablation_omega}
\end{table}

\begin{figure*}[!tb]
	\centering
	\subfigure{
		\begin{minipage}[t]{0.22\linewidth}
			\centering
			\includegraphics[width=1.\linewidth]{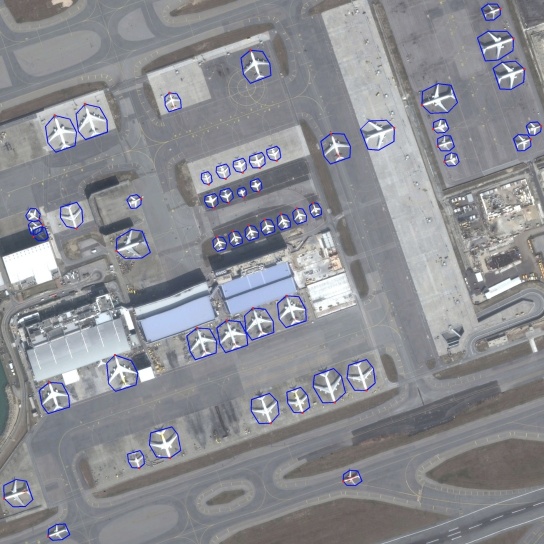}
			%\caption{fig1}
		\end{minipage}%
		\label{fig:ohd-sjtu-airplane-1}
	}
	\subfigure{
		\begin{minipage}[t]{0.22\linewidth}
			\centering
			\includegraphics[width=1.\linewidth]{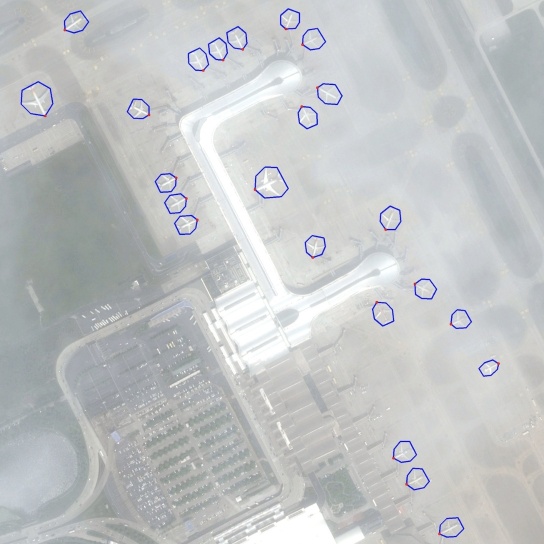}
			%\caption{fig2}
		\end{minipage}%
		\label{fig:ohd-sjtu-airplane-3}
	}
	\subfigure{
		\begin{minipage}[t]{0.22\linewidth}
			\centering
			\includegraphics[width=1.\linewidth]{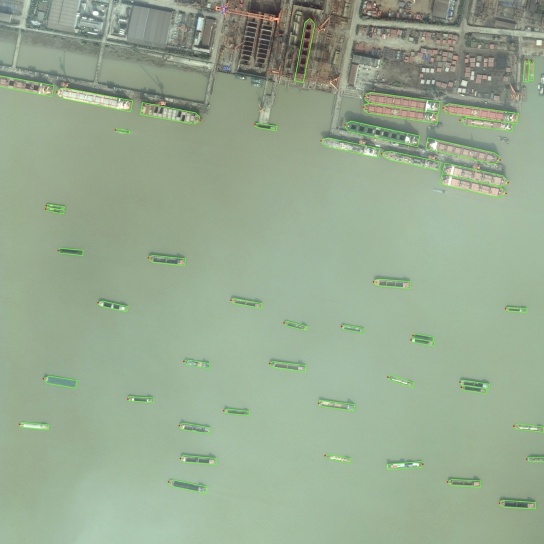}
			%\caption{fig2}
		\end{minipage}
		\label{fig:ohd-sjtu-ship-1}
	}
	\subfigure{
		\begin{minipage}[t]{0.22\linewidth}
			\centering
			\includegraphics[width=1.\linewidth]{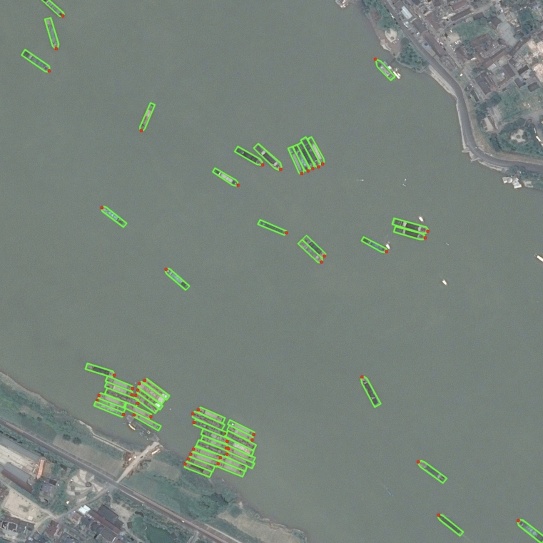}
			%\caption{fig2}
		\end{minipage}
		\label{fig:ohd-sjtu-ship-2}
	}
	\centering
	\vspace{-8pt}
	\caption{Samples of annotated subimages in our collected OHD-SJTU-S. Each object is labeled by an arbitrary quadrilateral, and the first marked point is the head position of the object to facilitate head prediction.}
	\label{fig:ohd-sjtu-dataset}
\end{figure*}

\begin{figure*}[tb!]
	\centering
	\subfigure[radius=0]{
		\begin{minipage}[t]{0.18\linewidth}
			\centering
			\includegraphics[width=1.\linewidth, height=4.2cm]{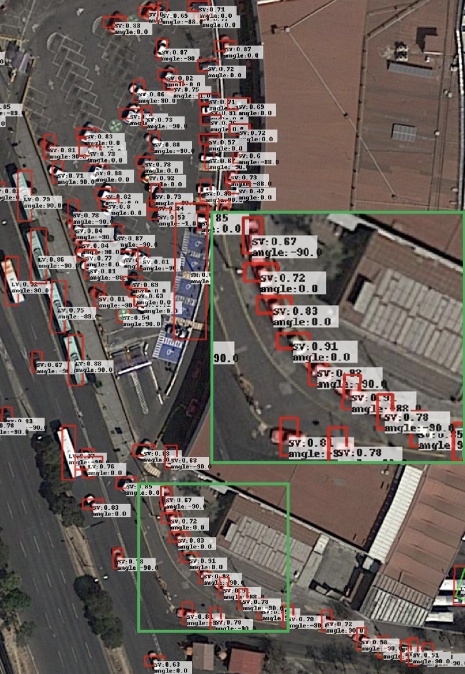}
			%\caption{fig1}
		\end{minipage}%
		\label{fig:r=0}
	}
	\subfigure[radius=2]{
		\begin{minipage}[t]{0.18\linewidth}
			\centering
			\includegraphics[width=1.\linewidth, height=4.2cm]{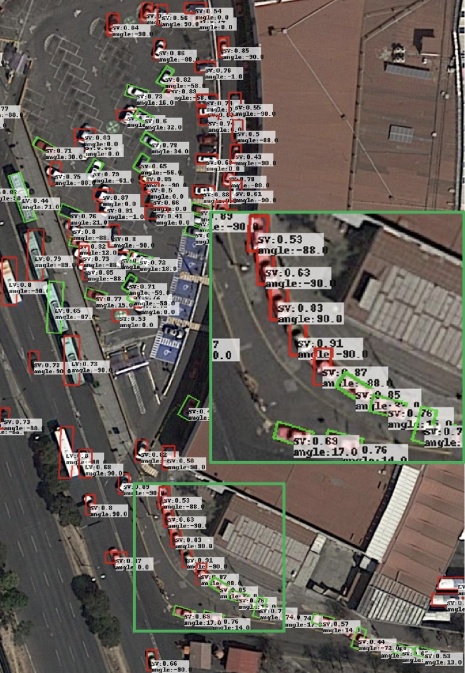}
			%\caption{fig2}
		\end{minipage}%
		\label{fig:r=2}
	}
	\subfigure[radius=4]{
		\begin{minipage}[t]{0.18\linewidth}
			\centering
			\includegraphics[width=1.\linewidth, height=4.2cm]{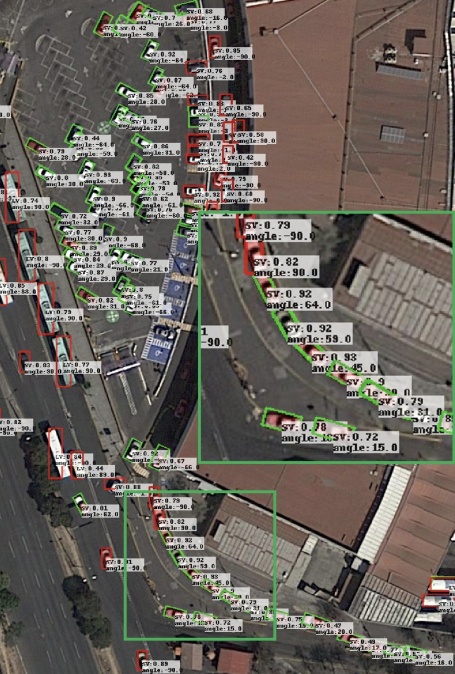}
			%\caption{fig2}
		\end{minipage}
		\label{fig:r=4}
	}
	\subfigure[radius=6]{
		\begin{minipage}[t]{0.18\linewidth}
			\centering
			\includegraphics[width=1.\linewidth, height=4.2cm]{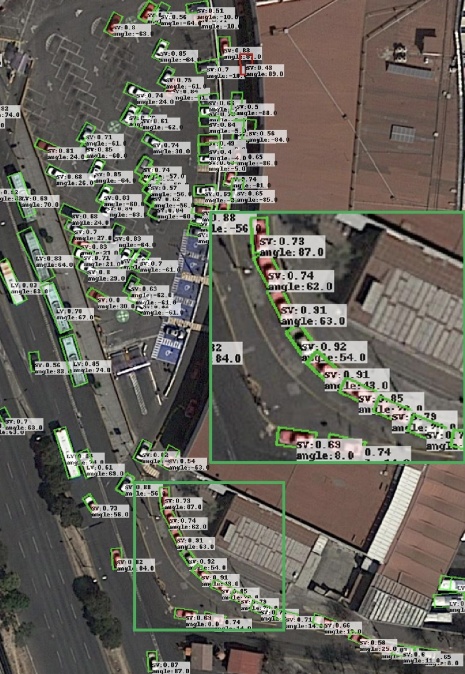}
			%\caption{fig1}
		\end{minipage}%
		\label{fig:r=6}
	}
	\subfigure[radius=8]{
		\begin{minipage}[t]{0.18\linewidth}
			\centering
			\includegraphics[width=1.\linewidth, height=4.2cm]{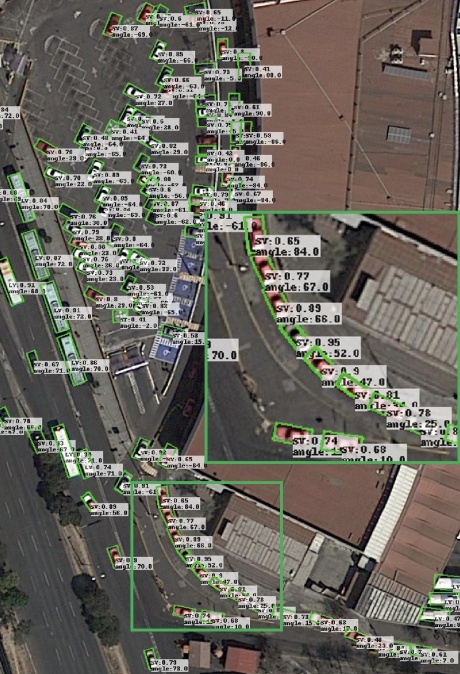}
			%\caption{fig2}
		\end{minipage}%
		\label{fig:r=8}
	} 
	\centering
	\caption{Visualization of detection results (RetinaNet-H CSL-Based) under different window function radius. The red bounding box indicates that no orientation and scale information has been learned, and the green bounding box is the correct detection result.}
	\label{fig:radius_vis}
	\vspace{-10pt}
\end{figure*}

\textbf{OHD-SJTU}
is our newly collected and public dataset for rotation detection and object heading detection. OHD-SJTU contains two different scale datasets, called OHD-SJTU-S and OHD-SJTU-L. OHD-SJTU-S is collected publicly from Google Earth with 43 large scene images sized $10, 000 \times 10, 000$ pixels and $16, 000 \times 16, 000$. It contains two object categories (ship and plane) and 4,125 instances (3,343 ships and 782 planes). Each object is labeled by an arbitrary quadrilateral, and the first marked point is the head position of the object to facilitate head prediction. We randomly select 30 original images as the training and validation set, and 13 images as the testing set. Figure \ref{fig:ohd-sjtu-dataset} shows some samples of annotated subimages in OHD-SJTU-S. The scenes cover a decent variety of road scenes and typical: cloud occlusion, seamless dense arrangement, strong changes in illumination/exposure, mixed sea and land scenes and large number of interfering objects. In contrast, OHD-SJTU-L adds more categories and instances, such as small vehicle, large vehicle, harbor, and helicopter. The additional data comes from DOTA-v1.0, but we reprocess the annotations and add the annotations of the object head. According to statistics, OHD-SJTU-L contains six object categories and 113,435 instances.
The statistical details are shown in Table \ref{tab:statistics}.
Compared with the AP$_{50}$ used by DOTA as the evaluation indicator, OHD-SJTU uses a more stringent AP$_{50:95}$ to measure the performance of the method, which poses a further challenge to the high accuracy of the detector. We divide the training and validation images into $ 600 \times 600 $ subimages with an overlap of 150 pixels and scale it to $ 800 \times 800 $. In the process of cropping the image with the sliding window, objects whose center point is in the subimage are kept. 

\begin{table}[!tb]
	\centering
	\caption{Comparison of detection results under different angle discretization granularities denoted by $\omega$ on the DOTA-v1.0 validation set.}
	\resizebox{0.48\textwidth}{!}{
		\begin{tabular}{c|c|ccccc|cccc}
			\toprule
			Method & $\omega$ & BR & SV & LV & SH & HA & 5-mAP$_{50}$ & mAP$_{50}$ & mAP$_{75}$ & mAP$_{50:95}$ \\
			\hline
			% \multirow{1}{*}{\shortstack{Reg-90}} & - & 39.01 & 53.92 & 51.05 & 75.32 & 61.21 & 56.10 & 64.70 & 32.50 & 34.50\\
			\multirow{1}{*}{\shortstack{Reg}} & - & 34.52 & 51.42 & 50.32 & 73.37 & 55.93 & 53.12 & 62.21 & 26.07 & 31.49\\
			\hline
			\multirow{1}{*}{\shortstack{CSL}} & 180/180 & 35.94 & 53.42 & 61.06 & 81.81 & 62.14 & 58.87 & 64.40 & 32.58 & 35.04\\
			\hline
            \multirow{8}{*}{\shortstack{BCL}}
            & 180/4 & 30.74 & 40.54 & 50.98 & 72.07 & 59.54 & 50.77 & 62.38 & 24.88 & 31.01 \\
            & 180/8 & 36.65 & 52.58 & 60.46 & 82.24 & 61.60 & 58.71 & \textbf{66.17} & 33.14 & 35.77 \\
            & 180/32 & \textbf{39.83} & 54.41 & 60.62 & 80.81 & 60.32 & \textbf{59.20} & 65.93 & \textbf{35.66} & \textbf{36.71} \\
            & 180/64 & 38.22 & \textbf{54.70} & 60.16 & 80.75 & 60.11 & 58.79 & 65.00 & 34.31 & 36.00 \\
            & 180/128 & 36.76 & 53.73 & \textbf{61.35} & \textbf{82.52} & 58.42 & 58.56 & 65.14 & 34.28 & 35.69 \\
            & 180/180 & 37.42 & 53.72 & 58.70 & 80.73 & \textbf{63.31} & 58.78 & 65.83 & 33.94 & 36.35 \\
            & 180/256 & 37.66 & 53.83 & 60.66 & 80.43 & 60.74 & 58.66 & 64.97 & 33.52 & 35.21 \\
            & 180/512 & 37.93 & 53.85 & 58.52 & 80.04 & 60.87 & 58.24 & 64.88 & 33.09 & 34.99 \\
            \hline
            \multirow{7}{*}{\shortstack{GCL}}
            & 180/4 & 30.90 & 41.20 & 48.30 & 72.93 & 60.16 & 50.70 & 62.98 & 23.83 & 30.81 \\
            & 180/8 & 36.88 & 51.10 & 59.81 & 82.40 & 61.57 & 58.35 & 65.23 & 33.92 & 35.29 \\
            & 180/32 & 38.04 & \textbf{54.77} & 60.88 & \textbf{82.75} & 61.24 & \textbf{59.54} & 65.11 & \textbf{34.67} & 36.15 \\
            & 180/64 & \textbf{38.05} & 54.36 & 60.59 & 81.84 & 60.39 & 59.05 & 64.78 & 33.23 & 35.67 \\
            & 180/128 & 37.74 & 54.36 & 59.43 & 81.15 & 60.51 & 58.64 & \textbf{66.13} & 33.65 & \textbf{36.34} \\
            % & 180/180 \\
            & 180/256 & 35.81 & 53.78 & 58.35 & 81.45 & 59.84 & 57.85 & 64.87 & 33.77 & 35.97 \\
            & 180/512 & 37.99 & 54.23 & \textbf{61.61} & 80.84 & \textbf{62.13} & 59.36 & 64.34 & 34.08 & 35.92 \\

			\bottomrule
	\end{tabular}}
	\label{table:ablation_omega_dcl}
\end{table}

All the used datasets are trained by 20 epochs in total, and the learning rate is reduced tenfold at 12 epochs and 16 epochs, respectively. The initial learning rates for RetinaNet and FPN are 5e-4 and 1e-3 respectively. The number of image iterations per epoch for DOTA-v1.0, DOTA-v1.5, DOTA-v2.0, ICDAR2015, MLT, HRSC2016, FDDB, OHD-SJTU-S and OHD-SJTU-L are 54k, 64k, 80k, 10k, 10k, 5k, 4k, 5k and 10k respectively, and doubled if data augmentation and multi-scale training are used. 

\subsection{Ablation Study}\label{sec:ablation_study}
\noindent \textbf{Comparison of classification and regression.}
Table \ref{tab:angle_pred} compares four different angle prediction methods on DOTA-v1.0 validation set, two of which are based on regression and the other two are via classification. Among them, the indirect regression (\textbf{Reg.$^*$}) is a relatively simple way to eliminate boundary problem, so it has a better performance than the direct regression (\textbf{Reg.}). In contrast, the classification-based prediction methods CSL, BCL and GCL outperform, achieving 35.04\%, 36.71\% and 36.34\% on the DOTA-v1.0 validation set, respectively.

\begin{figure*}[tb!]
	\centering
	\subfigure[$\omega_{csl}$=1]{
		\begin{minipage}[t]{0.18\linewidth}
			\centering
			\includegraphics[width=1.\linewidth]{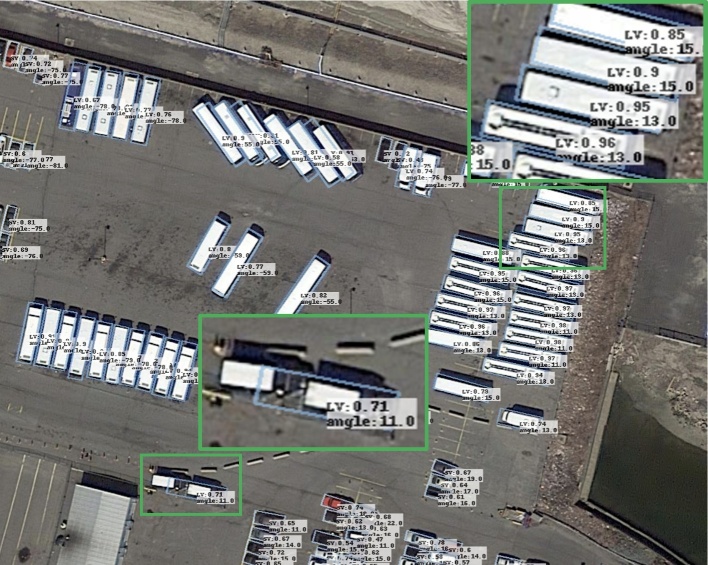}
			%\caption{fig1}
		\end{minipage}%
		\label{fig:P1925_w=1}
	}
	\subfigure[$\omega_{csl}$=3]{
		\begin{minipage}[t]{0.18\linewidth}
			\centering
			\includegraphics[width=1.\linewidth]{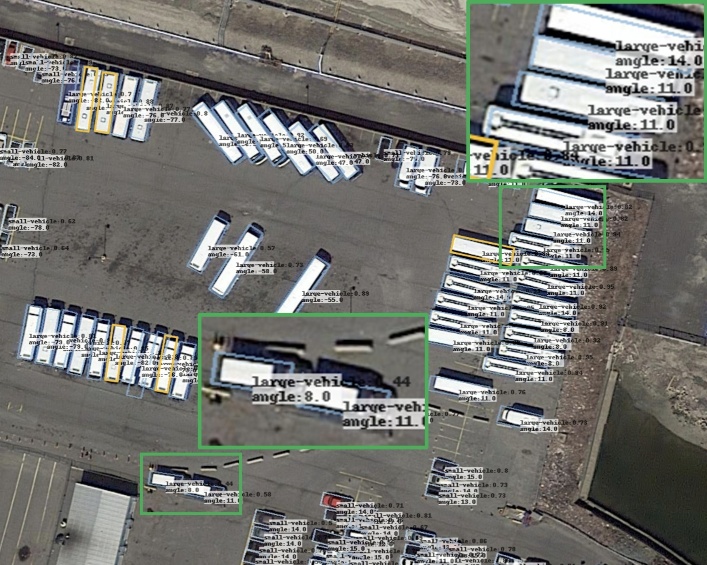}
			%\caption{fig2}
		\end{minipage}%
		\label{fig:P1925_w=3}
	}
	\subfigure[$\omega_{csl}$=10]{
		\begin{minipage}[t]{0.18\linewidth}
			\centering
			\includegraphics[width=1.\linewidth]{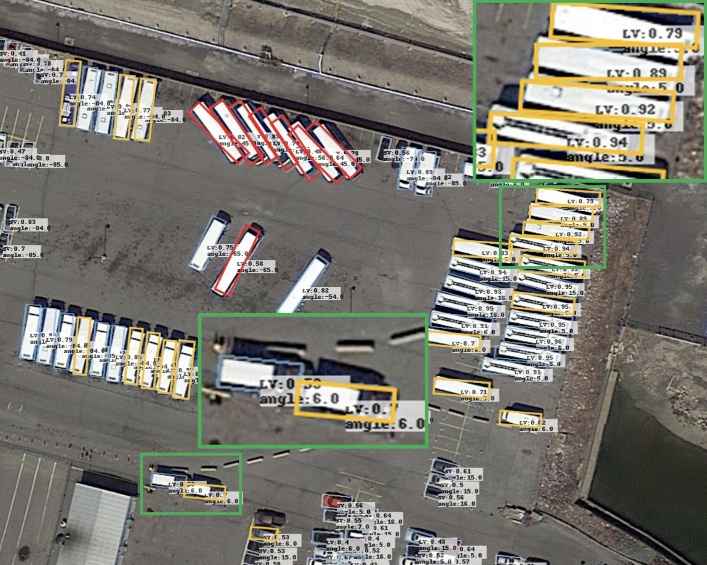}
			%\caption{fig2}
		\end{minipage}
		\label{fig:P1925_w=10}
	}
	\subfigure[$\omega_{csl}$=18]{
		\begin{minipage}[t]{0.18\linewidth}
			\centering
			\includegraphics[width=1.\linewidth]{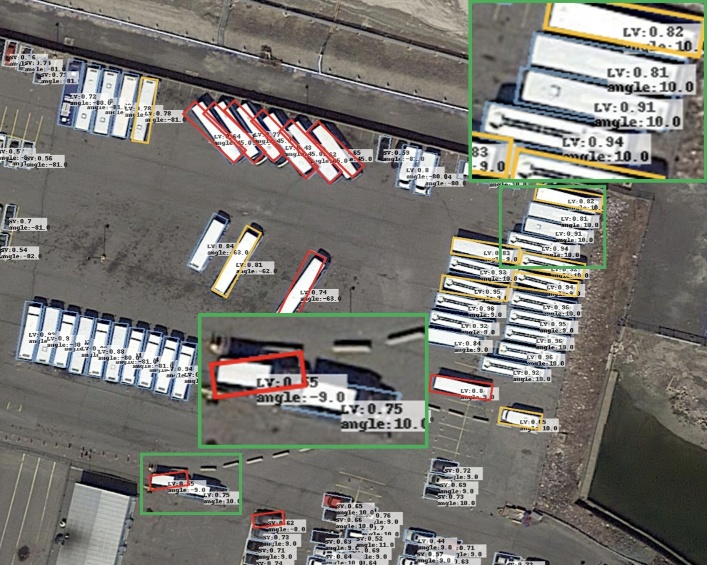}
			%\caption{fig1}
		\end{minipage}%
		\label{fig:P1925_w=18}
	}
	\subfigure[$\omega_{csl}$=30]{
		\begin{minipage}[t]{0.18\linewidth}
			\centering
			\includegraphics[width=1.\linewidth]{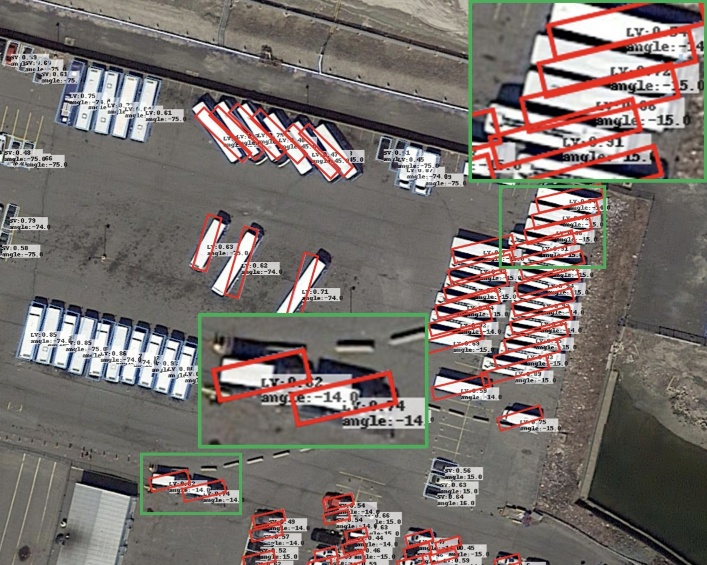}
			%\caption{fig2}
		\end{minipage}%
		\label{fig:P1925_w=30}
	} \\
	\vspace{-5pt}
	\subfigure[$\omega_{dcl}=180/4$]{
		\begin{minipage}[t]{0.23\linewidth}
			\centering
			\includegraphics[width=0.99\linewidth]{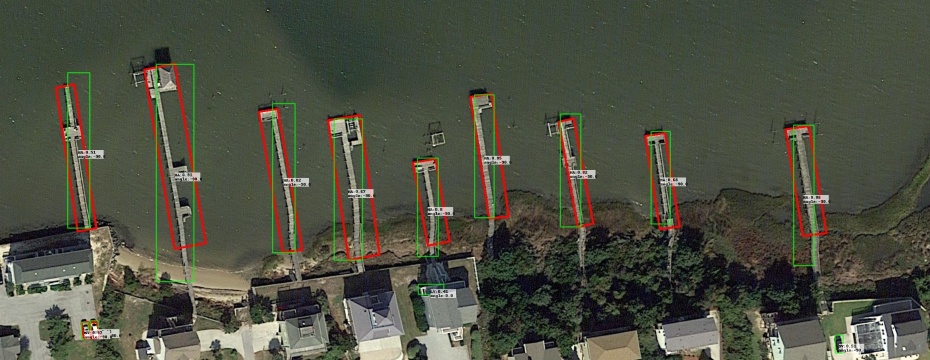}
		\end{minipage}%
		\label{fig:180/4}
	}
	\subfigure[$\omega_{dcl}=180/32$]{
		\begin{minipage}[t]{0.23\linewidth}
			\centering
			\includegraphics[width=0.99\linewidth]{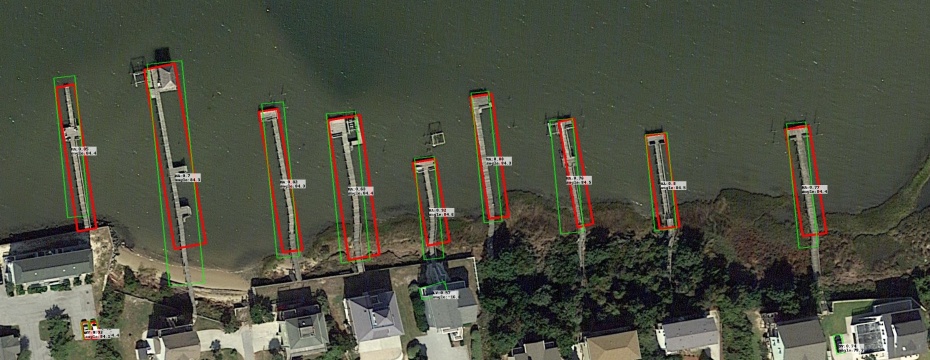}
		\end{minipage}%
		\label{fig:180/32}
	}
	\subfigure[$\omega_{dcl}=180/128$]{
		\begin{minipage}[t]{0.23\linewidth}
			\centering
			\includegraphics[width=0.99\linewidth]{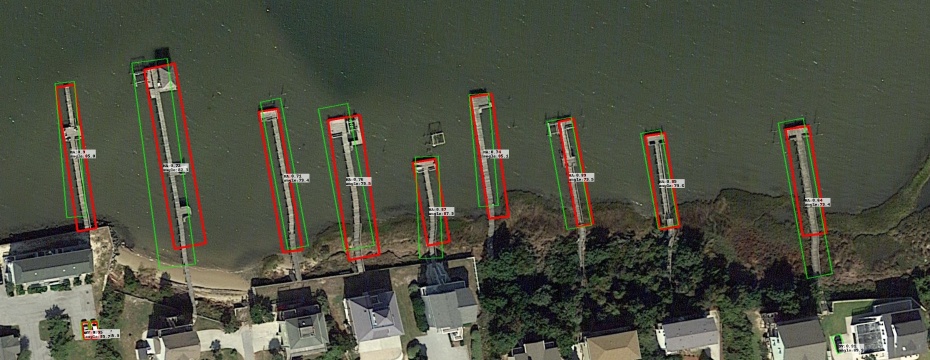}
		\end{minipage}
		\label{fig:180/128}
	}
	\subfigure[$\omega_{dcl}=180/256$]{
		\begin{minipage}[t]{0.23\linewidth}
			\centering
			\includegraphics[width=0.99\linewidth]{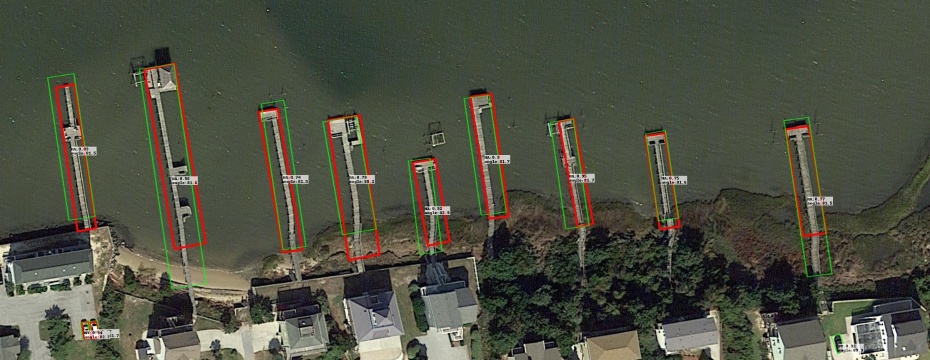}
		\end{minipage}%
		\label{fig:180/256}
	}
	\centering
	\caption{Visualization of detection results (RetinaNet-Based) under different angle discrete granularity $\omega$. For CSL, The red bounding box indicates that there is a large angle prediction error, and the orange bounding box indicates an acceptable angle prediction error. For DCL, the red and green box indicate ground truth and prediction.}
	\label{fig:omega_vis}
	\vspace{-5pt}
\end{figure*}

\begin{figure*}[!tb]
	\centering
	\subfigure[RetinaNet]{
		\begin{minipage}[t]{0.31\linewidth}
			\centering
			\includegraphics[width=0.98\linewidth]{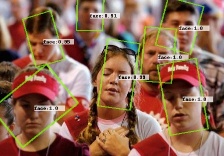}
		\end{minipage}%
		\label{fig:retinanet_fddb_P2170}
	}
	\subfigure[CSL]{
		\begin{minipage}[t]{0.31\linewidth}
			\centering
			\includegraphics[width=0.98\linewidth]{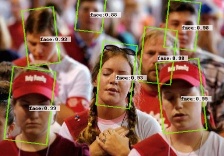}
		\end{minipage}%
		\label{fig:csl_fddb_P2170}
	}
	\subfigure[DCL]{
		\begin{minipage}[t]{0.31\linewidth}
			\centering
			\includegraphics[width=0.98\linewidth]{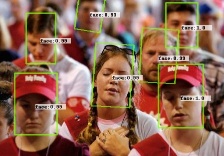}
		\end{minipage}%
		\label{fig:bcl_fddb_P2170}
	}
	\centering
	\caption{Visual comparison between classification-based and regression-based protocols on the FDDB dataset.}
	\label{fig:vis_fddb}
\end{figure*}

\begin{table*}[tb!]
	\centering
	\caption{Comparison between classification-based and regression-based protocols on the DOTA-v1.0 test set. `H' and `R' represent the horizontal and rotating anchors, respectively.}
	\resizebox{0.98\textwidth}{!}{
		\begin{tabular}{c|c|c|c|c|c|c|c|c|c|c|c|c}
			\toprule
			Baseline & Angle Range & Angle Pred. & PoA & EoE &  Label Mode
			& BR & SV & LV & SH & HA & 5-mAP & mAP \\
			\hline
			\multirow{8}{*}{RetinaNet-H}
			& \multirow{3}{*}{$[-90^\circ,0^\circ)$} & Reg. ($\Delta \theta$) & $\checkmark$ & $\checkmark$ & - & 41.15 & 53.75 & 48.30 & 55.92 & 55.77 & 50.98 & 63.18 \\
			& & Cls.: CSL & & $\checkmark$ & Gaussian & 41.03 & 59.63 & 52.57 & 64.56 & 54.64 & 54.49 \textbf{(+3.51)} & 63.51 \textbf{(+0.33)} \\
			& & Cls.: GCL & & $\checkmark$ & - & 42.49 & 62.38 & 49.13 & 69.06 & 56.40 & 55.89 \textbf{(+4.09)} & 64.94 \textbf{(+1.76)} \\
			\cline{2-13}
			& \multirow{5}{*}{$[-90^\circ,90^\circ)$} & Reg. ($\Delta \theta$) & $\checkmark$ &  & \- & 38.31 & 60.48 & 49.77 & 68.29 & 51.28 & 53.63 & 64.17 \\
			& & Reg.$^*$ ($\sin{\theta}, \cos{\theta}$) & & & - & 41.52 & 63.94 & 44.95 & 71.18 & 53.22 & 54.96 \textbf{(+1.33)} & 65.78 \textbf{(+1.61)}\\
			& & Cls.: CSL & & & Gaussian & 42.25 & 68.28 & 54.51 & 72.85 & 53.10 & 58.20 \textbf{(+4.57)} & 67.38 \textbf{(+3.21)} \\
			& & Cls.: GCL & & & - & 39.78 & 67.20 & 56.02 & 74.10 & 53.82 & 58.18 \textbf{(+4.55)} & 67.02 \textbf{(+2.85)} \\
			& & Cls.: BCL & & & - & 41.40 & 65.82 & 56.27 & 73.80 & 54.30 & 58.32 \textbf{(+4.69)} & 67.39 \textbf{(+3.22)} \\
			\hline
			\multirow{2}{*}{RetinaNet-R}
			&  \multirow{2}{*}{$[-90^\circ,0^\circ)$} & Reg. ($\Delta \theta$) & $\checkmark$ & $\checkmark$ & - & 32.27 & 64.64 & 71.01 & 68.62 & 53.52 & 58.01 & 62.76 \\
			& & Cls.: CSL & & $\checkmark$ & Gaussian & 35.14 & 63.21 & 73.92 & 69.49 & 55.53 & 59.46 \textbf{(+1.45)} & 65.45 \textbf{(+2.69)} \\
			\hline
			\multirow{2}{*}{R$^3$Det} & \multirow{2}{*}{$[-90^\circ,0^\circ)$} & Reg. ($\Delta \theta$) & $\checkmark$ & $\checkmark$ & - & 44.15 & 75.09 & 72.88 & 86.04 & 61.01 & 67.83 & 70.66\\
			& & Cls.: BCL & & $\checkmark$ & Gaussian & 46.84 & 74.87 & 74.96 & 85.70 & 57.72 & 68.02 \textbf{(+0.19)} & 71.21 \textbf{(+0.55)}\\
			\hline
			\multirow{5}{*}{FPN-H}
			&  \multirow{2}{*}{$[-90^\circ,0^\circ)$} & Reg. ($\Delta \theta$) & $\checkmark$ & $\checkmark$ & - & 44.78 & 70.25 & 71.13 & 68.80 & 54.27 & 61.85 & 68.25   \\
			& & Cls.: CSL & & $\checkmark$ & Gaussian & 45.46 & 70.22 & 71.96 & 76.06 & 54.84 & 63.71 \textbf{(+1.86)} & 69.02 \textbf{(+0.77)}  \\
			\cline{2-13}
			& \multirow{3}{*}{$[-90^\circ,90^\circ)$} & Reg. ($\Delta \theta$) & $\checkmark$ & & - & 45.88 & 69.37 & 72.06 & 72.96 & 62.31 & 64.52 & 69.45 \\
			& & Cls.: CSL & & & Gaussian & 47.90 & 69.66 & 74.30 & 77.06 & 64.59 & 66.70 \textbf{(+2.18)} & 70.92 \textbf{(+1.47)} \\
			& & Cls.: GCL & & & - & 47.56 & 69.81 & 74.03 & 76.56 & 64.29 & 66.45 \textbf{(+1.93)} & 70.83 \textbf{(+1.38)} \\
			\bottomrule
	\end{tabular}}
	\label{table:ablation_csl_gcl}
% 	\vspace{-5pt}
\end{table*}

\noindent \textbf{Comparison of four window functions in CSL method.} Table \ref{table:ablation_window_func} shows the performance comparison of the four window functions on the DOTA-v1.0 dataset. It also details the accuracy of the five categories with larger aspect ratio and more border cases in the dataset. We believe that these categories can better reflect the advantages of our method. In general, the Gaussian window function performs best, while the pulse function performs worst because it has not learned any orientation and scale information. Figures \ref{fig:csl-pulse}-\ref{fig:csl-gauss-180} show the visualization of the four window functions. According to Figure \ref{fig:csl-gauss-180} and Figure \ref{fig:csl-gauss-90}, the $180^\circ$-CSL-based protocol obviously has better boundary prediction due to the EoE problem still exists in the $90^\circ$-CSL-based protocol. Figure \ref{fig:win_func_vis} shows the consistent results with the those in Table \ref{table:ablation_window_func}.

\noindent \textbf{Suitable window radius in CSL method.}  The Gaussian window form has shown best performance, while here we study the effect of radius of the window function. When the radius is too small, the window function tends to a pulse function. Conversely, the discrimination of all predictable results becomes smaller when the radius is too large. Therefore, we choose a suitable radius range from 0 to 8, Table \ref{table:ablation_radius} shows the performance of the two detectors in this range. Although both detectors achieve the best performance with a radius of 6, the single-stage detection method is more sensitive to radius. We speculate that the instance-level feature extraction capability (like RoI Pooling \cite{girshick2015fast} and RoI Align \cite{he2017mask}) in the two-stage detector is stronger than the image-level in the single-stage detector. Therefore, the two-stage detection method can distinguish the difference between the two approaching angles. Figure \ref{fig:radius_vis} compares visualizations using different window radius. When the radius is 0, the detector cannot learn any orientation and scale information, which is consistent with the performance of the pulse function above. As the radius becomes larger and more optimal, the detector can learn the angle in any direction. Please refer to the enlarged part of the figure.

\begin{figure*}[tb!]
	\centering
	\subfigure[bin=90]{
		\begin{minipage}[t]{0.23\linewidth}
			\centering
			\includegraphics[width=1.\linewidth]{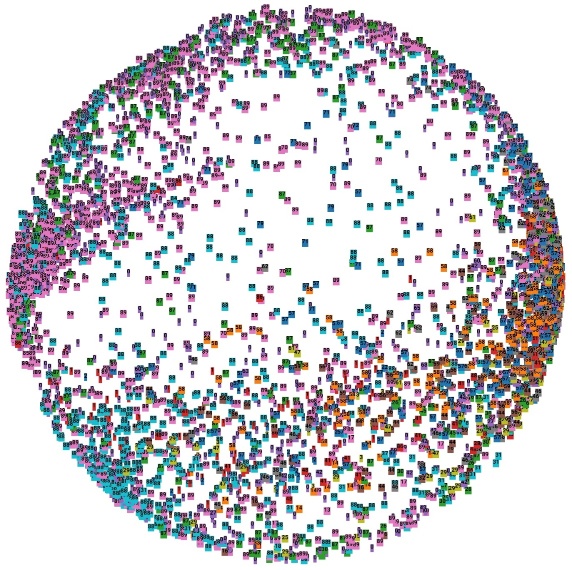}\vspace{0.15cm}
			\centering
			\includegraphics[width=1.\linewidth]{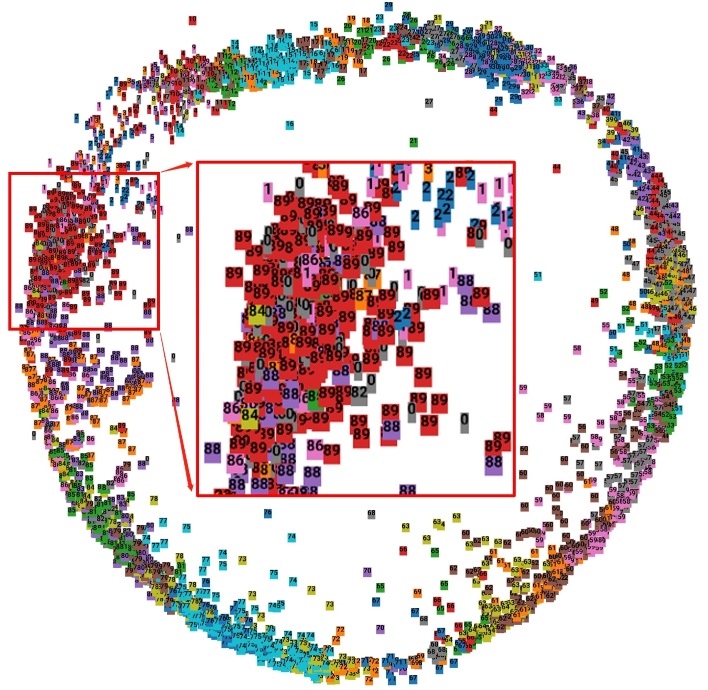}\vspace{0.15cm}
			%			\centering
			%			\includegraphics[width=0.95\linewidth]{figure_ijcv/pca-1-3d.jpg}
		\end{minipage}%
		\label{fig:pca-1}
	}
	\subfigure[bin=15]{
		\begin{minipage}[t]{0.23\linewidth}
			\centering
			\includegraphics[width=1.\linewidth]{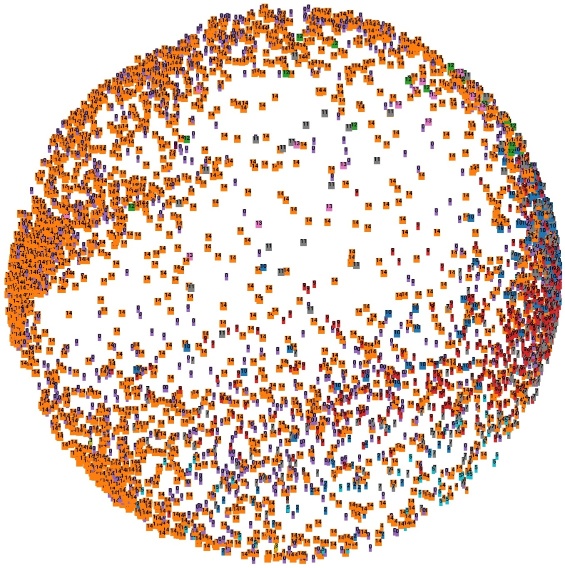}\vspace{0.15cm}
			\centering
			\includegraphics[width=1.\linewidth]{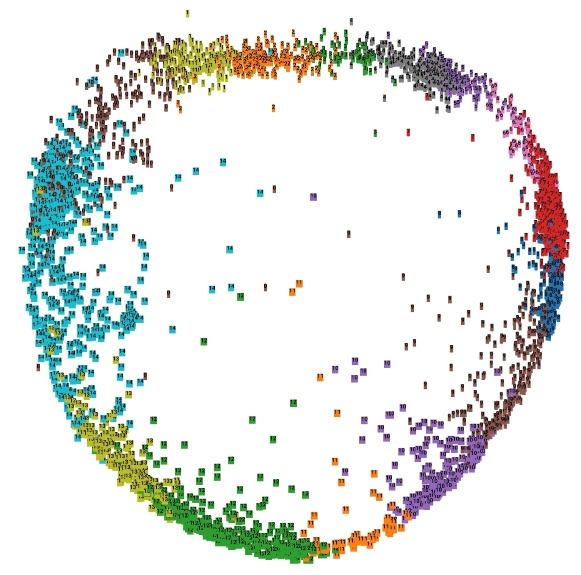}\vspace{0.15cm}
			%			\centering
			%			\includegraphics[width=0.95\linewidth]{figure_ijcv/pca-6-3d.jpg}
		\end{minipage}%
		\label{fig:pca-6}
	}
	\subfigure[bin=9]{
		\begin{minipage}[t]{0.23\linewidth}
			\centering
			\includegraphics[width=1.\linewidth]{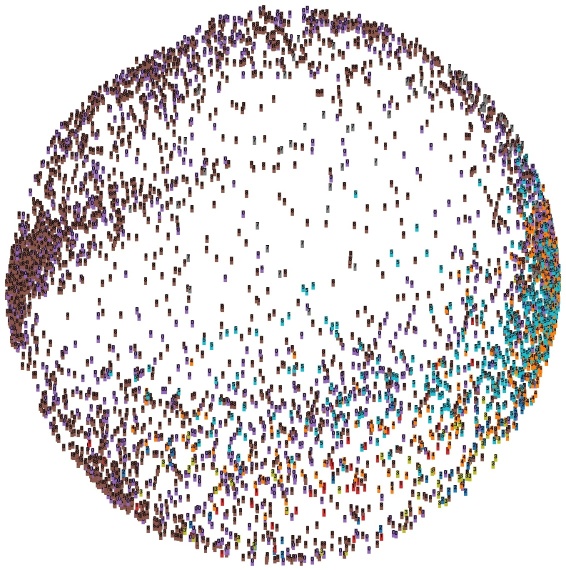}\vspace{0.15cm}
			\centering
			\includegraphics[width=1.\linewidth]{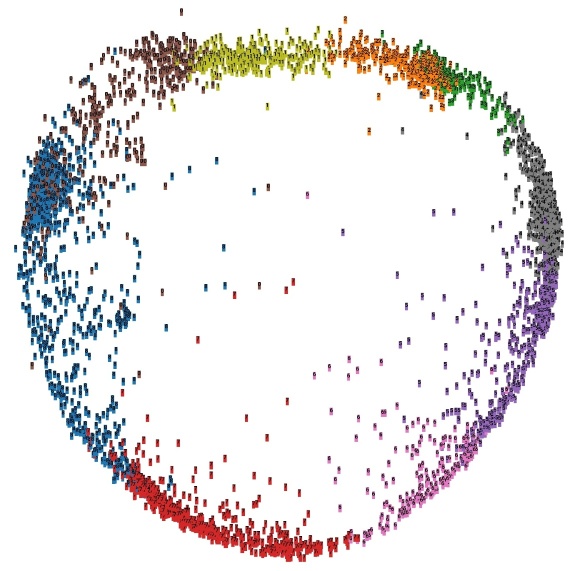}\vspace{0.15cm}
			%			\centering
			%			\includegraphics[width=0.95\linewidth]{figure_ijcv/pca-10-3d.jpg}
		\end{minipage}%
		\label{fig:pca-10}
	}
	\subfigure[bin=6]{
		\begin{minipage}[t]{0.23\linewidth}
			\centering
			\includegraphics[width=1.\linewidth]{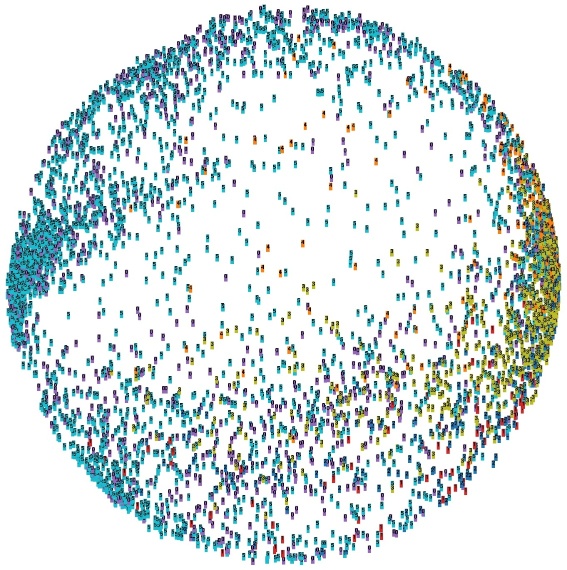}\vspace{0.15cm}
			\centering
			\includegraphics[width=1.\linewidth]{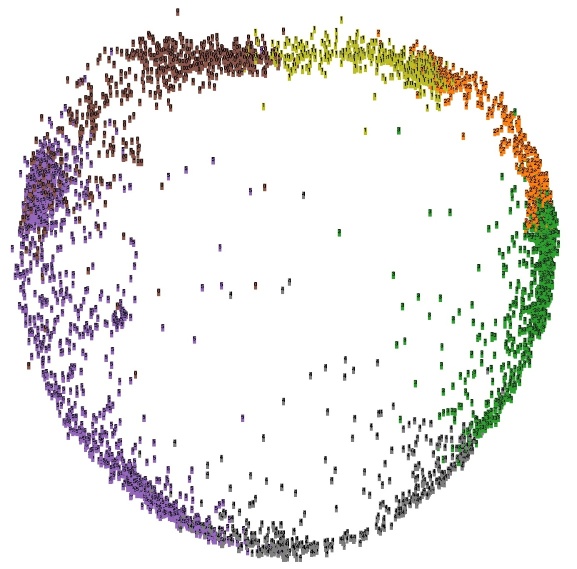}\vspace{0.15cm}
			%			\centering
			%			\includegraphics[width=0.95\linewidth]{figure_ijcv/pca-15-3d.jpg}
		\end{minipage}%
		\label{fig:pca-15}
	}
	\centering
	\caption{Angular feature visualization of the 90-CSL-FPN detector on the DOTA-v1.0 dataset. First, we divide the entire angular range into several bins, and bins are different between columns. The two rows show two-dimensional feature visualizations of pulse and Gaussian function, respectively. Each point represents an RoI of the test set with an index of the bin it belongs to.}
	\label{fig:vis_feature}
\end{figure*}

\begin{figure}[!tb]
	\centering
	\subfigure[$\omega=180/4$]{
		\begin{minipage}[t]{0.45\linewidth}
			\centering
			\includegraphics[width=0.98\linewidth]{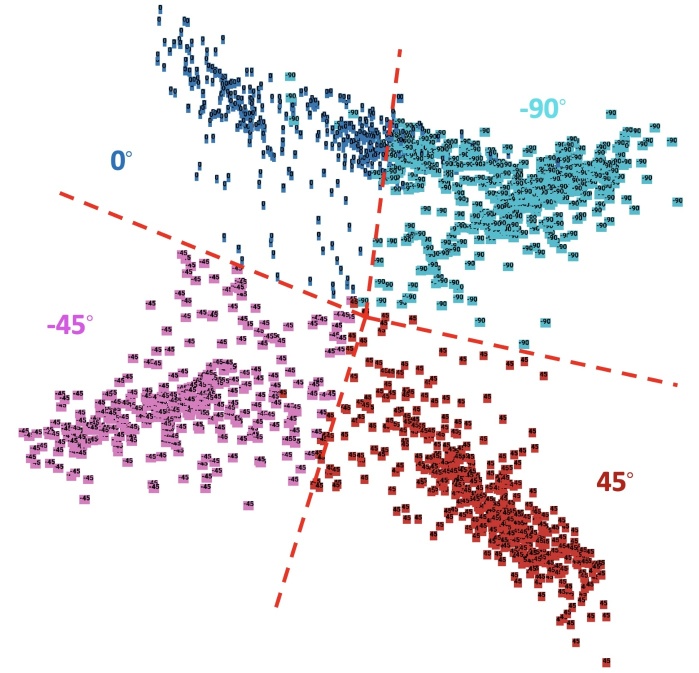}
		\end{minipage}%
		\label{fig:feature_vis_4}
	}
	\subfigure[$\omega=180/8$]{
		\begin{minipage}[t]{0.45\linewidth}
			\centering
			\includegraphics[width=0.98\linewidth]{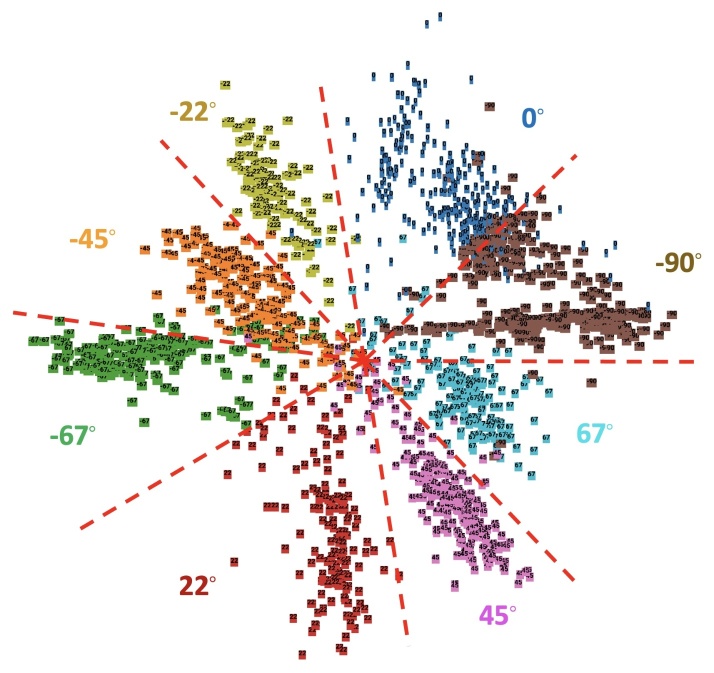}
		\end{minipage}%
		\label{fig:feature_vis_8}
	}
	\centering
	\caption{Angular feature visualization of the RetinaNet-DCL. The red dotted lines divide the categories.}
	\label{fig:feature_vis_dcl}
\end{figure}

\noindent \textbf{Angle discretization granularity $\omega$.}
In general, the smaller the angle discrete granularity $\omega$, the more accurate the angle predicted by the model, as shown in Figure \ref{fig:omega_vis}. However, considering that the criterion for judging the object to be detected is that IoU is greater than 0.5 (such as the DOTA-v1.0 dataset). Therefore, a proper $\omega$ can make the model have a certain degree of fault tolerance, and can achieve better detection performance. It can be seen from Table \ref{table:ablation_omega} that when the discrete granularity is 10, the CSL-based protocol can achieve the highest performance on the DOTA-v1.0 dataset, and when $\omega$ is 30, the excessive angular prediction error makes the performance of the model drop sharply. Discrete granularity $\omega$ can be approximated as a CSL technique with a rectangular window function, which has a certain tolerance in the divided angle interval. The difference between them is that CSL smooths between adjacent angle intervals.

Similar conclusions can still be obtained in the DCL method, as shown in Table \ref{table:ablation_omega} and Table \ref{table:ablation_omega_dcl}. In general, the smaller $\omega$, the higher theoretical upper bound of the model's performance. However, the decrease of $\omega$ will lead to an increase in the number of angle categories, which poses a challenge to the angle classification performance of the model. Therefore, we need to explore the impact of $\omega$ on the detection performance under different IoU thresholds, and find a suitable range of $\omega$. In order to get the performance indicators under different IoU threshold, we conduct experiments on the DOTA-v1.0 validation set, and the number of image iterations per epoch is 40k. According to Table \ref{table:ablation_omega_dcl}, when the number of angle categories is between 32 and 128, the performance of the model reaches its peak. If the number of categories is too small, the theoretical accuracy loss is too large, resulting in a sharp drop in performance; if the number of categories is too large, the angle classification network of the model cannot be effectively processed and the performance will decrease slightly. Figure \ref{fig:omega_vis} shows the comparison of angle estimates under different $\omega$. Compared with CSL, DCL can set a smaller $w$ (e.g. $w < 1$) without too much parameter overhead, and no need to adjust the window function radius at the same time.

\noindent \textbf{Redundant invalid coding.}
To make each code have a corresponding different angle value, the number of categories must be a power of 2 in the DCL-based method. However, this is not required. When we only set 180 categories, about 76 codings are invalid, but BCL-based method can still achieve good performance, at 36.35\% as shown in Table \ref{table:ablation_omega_dcl}. We also artificially increase the length based on the theoretical shortest code length to increase the proportion of invalid codes, and the performance is only slightly reduced.

\begin{table*}[tb!]
	\centering
	\caption{Comparison between classification-based and regression-based protocols on the text dataset ICDAR2015, MLT, aerial dataset HRSC2016, and face dataset FDDB. Note 2007 and 2012 in bracket means using the 2007 and 2012 evaluation metric, respectively. Except for FDDB's baseline is RetinaNet \cite{lin2017focal}, the others are FPN \cite{lin2017feature}.}
	\resizebox{0.98\textwidth}{!}{
		\begin{tabular}{c|ccc|ccc|cc|ccc}
			\toprule
			\multirow{2}{*}{Method}&\multicolumn{3}{c|}{ICDAR2015}&\multicolumn{3}{c|}{MLT}&\multicolumn{2}{c|}{HRSC2016} &\multicolumn{2}{c}{FDDB} \\
			\cline{2-4} \cline{5-7} \cline{8-9} \cline{10-11}
			& Recall & Precision & Hmean & Recall & Precision & Hmean  & mAP (2007) & mAP (2012) & AP$_{50}$ (2012) & AP$_{75}$ (2012) \\
			\hline
			baseline & 81.81 & 83.07 & 82.44 & 56.15 & 80.26 & 66.08 & 88.33 & 94.70 & 95.92 & 55.81 \\
			CSL & 83.00 & 84.30 & 83.65 \textbf{(+1.21)} & 56.72 & 80.77 & 66.64 \textbf{(+0.56)} & 89.62 \textbf{(+1.29)} & 96.10 \textbf{(+1.40)} & 96.64 \textbf{(+0.72)} & 73.22 \textbf{(+17.41)} \\
			GCL & 82.56 & 84.72 & 83.63 \textbf{(+1.19)} & 57.54 & 80.65 & 67.16 \textbf{(+1.08)} & 89.56 \textbf{(+1.23)} & 96.02 \textbf{(+1.32)} & 96.16 \textbf{(+0.24)} & 74.06 \textbf{(+18.25)} \\
			\bottomrule
		\end{tabular}
		\label{table:other_dataset_comparison}
	}
\end{table*}

\begin{table}[tb!]
    \centering
    \caption{Ablative study by accuracy (\%) of CSL and DCL on the OBB task of DOTA-v1.0/v1.5/v2.0.}
    \resizebox{0.48\textwidth}{!}{
    \begin{tabular}{c|c|c|c|c}
        \hline
        Method & Angle Pred. & DOTA-v1.0 & DOTA-v1.5 & DOTA-v2.0\\ 
        \hline
        \multirow{3}{*}{RetinaNet-H} & Reg. $(\Delta\theta)$ & 64.17 & 56.10 & 43.06 \\
        & Cls: CSL & 67.38 & 58.55 & 43.34 \\
        & Cls: BCL & \textbf{67.39} & \textbf{59.38} & \textbf{45.46} \\
        \hline
    \end{tabular}}
    \label{tab:dota1to2}
\end{table}

\begin{table}[tb!]
	\centering
	\caption{Ablation experiment of our proposed angle fine-tuning technique on the HRSC2016 dataset. The baseline is RetinaNet.}
	\resizebox{0.48\textwidth}{!}{
		\begin{tabular}{cc|cc|cc|cc}
			\toprule
			\multirow{2}{*}{Method} & \multirow{2}{*}{Fine-Tune} & \multicolumn{2}{c|}{$\omega_{csl}$=18, $\omega_{gcl}$=180/8} & \multicolumn{2}{c|}{$\omega_{csl}$=10, $\omega_{gcl}$=180/128} & \multicolumn{2}{c}{$\omega_{csl}$=1, $\omega_{gcl}$=180/256} \\
			\cline{3-8}
			& & AP$_{50}$ & AP$_{75}$ & AP$_{50}$ & AP$_{75}$ & AP$_{50}$ & AP$_{75}$ \\
			\hline
			\multirow{2}{*}{CSL} & & 46.72 & 3.68 & 74.07 & 16.89 & 81.17 & 44.75 \\
			& $\checkmark$ & 76.17 & 20.47 & 81.41 & 32.57 & 81.37 & 46.51 \\
			\hline
			\multirow{2}{*}{GCL} & & 57.34 & 5.36 & 77.77 & 30.35 & 76.60 & 34.20\\
			& $\checkmark$ & 66.70 & 28.27 & 78.07 & 32.62 & 76.56 & 34.60\\
			\bottomrule
	\end{tabular}}
	\label{table:ablation_angle_fine_tuning}
\end{table}

\noindent \textbf{Performance of CSL and DCL on other detectors.}
Four detectors in Table \ref{table:ablation_csl_gcl}, including RetinaNet-H, RetinaNet-R, R$^3$Det and FPN-H, are used to compare the performance differences among CSL-based, DCL-based and regression-based protocols. The former two are single-stage detectors, whose anchor format is different. 
One of the remaining two is a cascade multi-stage strategy based method and the other is a classic two-stage detection method.
It can be clearly seen that CSL and DCL have better detection ability for objects with large aspect ratios and more boundary conditions. It also should be noted that CSL and DCL are designed to solve the boundary problem, whose proportion in the entire dataset is relatively small, so the overall performance (mAP) is not as obvious as the five categories listed (5-mAP). Overall, the CSL-based and DCL-based rotation detection algorithms are indeed better baseline choices than the angle regression-based protocol.

\noindent \textbf{Performance of CSL and DCL on other datasets.}
To further verify that CSL-based and DCL-based protocols are better than baseline, we have also verified it in other datasets, including the text dataset ICDAR2015, MLT, and another remote sensing dataset HRSC2016. These three datasets are single-class object detection datasets, whose objects have a large aspect ratio. Although boundary conditions still account for a small proportion of these datasets, CSL and DCL still shows stronger performance advantage. As shown in Table \ref{table:other_dataset_comparison}, compared with the regression-based protocol, the CSL-based protocol is improved by 1.21\%, 0.56\%, and 1.29\% (1.4\%) respectively under the same experimental configuration. The same improvement is also reflected in the DCL-based protocol.
For face dataset FDDB (Figure \ref{fig:vis_fddb}), AP$_{50}$ cannot better reflect the advantages of the proposed technique due to the aspect ratio of face is small. In contrast, CSL/GCL shows a great performance improvement on AP$_{75}$, about 17.41\%/18.25\%.
For more challenging datasets (e.g. DOTA-v1.5, DOTA-v2.0 in Table \ref{tab:dota1to2}), CSL/DCL still has a steady improvement.
These experimental results provide strong support for demonstrating the versatility of the CSL-based and DCL-based protocols.

\noindent \textbf{Angle Fine-Tuning.}
Table \ref{table:ablation_angle_fine_tuning} compares the performance before and after using angle fine-tuning under different sizes of $\omega$ on the HRSC2016.
The low-precision indicator (mAP$_{50}$) has a certain tolerance for angle errors.
Take two objects with the same scale and the same center as an example, when their aspect ratio is 1:9, their IoU is still close to 0.7 when the angle deviation is $5^\circ$, which shows that mAP$_{50}$ cannot better reflect the advantages of the angle fine-tuning mechanism when $\omega$ is small.
When $\omega_{csl}=1$, $\omega_{gcl}=180/256$, the maximum angle that can be fine-tuned is only $0.5^\circ$ and $0.35^\circ$ according to Eq. \ref{eq:theta_error}, so the effectiveness of fine-tuning technique on mAP$_{50}$ is not significant in Table \ref{table:ablation_angle_fine_tuning}. In contrast, the high-precision indicator (mAP$_{75}$) well reflects the advantages of the angle fine-tuning technique, and the larger the $\omega$, the more significant its advantages. When $\omega$ becomes larger that the theoretical accuracy exceeds the tolerance of mAP$_{50}$, angle fine-tuning technique can effectively adjust the prediction angle to reduce the negative impact of theoretical angle errors and improve both mAP$_{50}$ and mAP$_{75}$.

\begin{table}[tb!]
	\centering
	\caption{Accuracy and speed on HRSC2016. Here (07) and (12) means using the 2007 and 2012 evaluation metric, respectively.}
	\resizebox{0.48\textwidth}{!}{
		\begin{tabular}{lccc}
			\toprule
			
			Method & Backbone & mAP (07) & mAP (12)\\
			
			\midrule
            R$^2$CNN \cite{jiang2017r2cnn} & ResNet101 & 73.07 & 79.73 \\
			RC1 \& RC2 \cite{liu2017high} & VGG16 & 75.70 & -- \\
			RRPN \cite{ma2018arbitrary} & ResNet101 & 79.08 & 85.64 \\
			R$^2$PN \cite{zhang2018toward}  & VGG16 & 79.60 & --  \\
			RetinaNet-H \cite{yang2021r3det} & ResNet101 & 82.89 & 89.27 \\
			RRD \cite{liao2018rotation} & VGG16  & 84.30 & --  \\
			RoI-Transformer \cite{ding2018learning} & ResNet101 & 86.20 & -- \\
			Gliding Vertex \cite{xu2020gliding} & ResNet101 & 88.20 & -- \\
			BBAVectors \cite{yi2021oriented} & ResNet101 & 88.60 & -- \\
			DRN \cite{pan2020dynamic} & Hourglass104 & -- & 92.70 \\
			CenterMap OBB \cite{wang2020learning} & ResNet50 & -- & 92.80 \\
			SBD \cite{liu2019omnidirectional} & ResNet50 & -- & 93.70 \\
			RetinaNet-R \cite{yang2021r3det} & ResNet101 & 89.18 & 95.21 \\
			R$^3$Det \cite{yang2021r3det} & ResNet101 & 89.26 & 96.01 \\
			\hline
            CSL & ResNet101 & \textbf{89.62} & 96.10 \\
            GCL & ResNet101 & 89.56 & 96.02\\
            BCL & ResNet101 & 89.46 & \textbf{96.41}\\
			\bottomrule
			
	\end{tabular}}
	\label{table:HRSC2016}
\end{table}

\begin{table}[tb!]
	\centering
	\caption{Detection accuracy on ICDAR2015. $^\dag$ and $^\ddag$ denote that the method uses external data and stronger pre-trained weight, respectively.}
	\resizebox{0.48\textwidth}{!}{
		\begin{tabular}{l|c|c|ccc}
			\toprule
			
			Method & Venue & Backbone & Precison & Recall & F-measure \\
			\hline
		    CTPN \cite{tian2016detecting} & ECCV'16 & VGG16 & 74.2 & 51.5 & 60.8\\
		    EAST$^\dag$ \cite{zhou2017east} & CVPR'17 & VGG16 & 83.5 & 73.4 & 78.2\\
		    DeepReg \cite{he2017deep} & ICCV'17 & VGG16 & 82.0 & 80.0 & 81.0\\
		    RRPN$^\dag$ \cite{ma2018arbitrary} & TMM'18 & VGG16 & 82.0 & 73.0 & 77.0 \\
		    PixelLink \cite{deng2018pixellink} & AAAI'18 & VGG16 & 82.9 & 81.7 & 82.3\\
		    PAN$^\ddag$ \cite{wang2019efficient} & ICCV'19 & ResNet18 & 82.9 & 77.8 & 80.3\\
		    TextField$^{\dag\ddag}$ \cite{xu2019textfield} & TIP'19 & VGG16 & 84.3 & 80.5 & 82.4 \\
		    TextDragon$^{\dag\ddag}$ \cite{feng2019textdragon} & ICCV'19 & VGG16 & 84.8 & 81.8 & 83.1 \\
		    DBNet(736)$^\ddag$ \cite{liao2020real} & AAAI'20 & ResNet18 & 86.8 & 78.4 & 82.3 \\
		    DBNet(1,152)$^\ddag$ \cite{liao2020real} & AAAI'20 & ResNet50 &  91.8 & 83.2 & 87.3 \\
		    PAN++$^\ddag$ \cite{wang2021pan++} & TPAMI'21 & ResNet18 & 86.7 & 78.4 & 82.3\\
		    PAN++$^{\dag\ddag}$ \cite{wang2021pan++} & TPAMI'21 & ResNet50 & 91.4 & 83.9 & \textbf{87.5}\\
		    PolarMask++$^\ddag$ \cite{xie2021polarmask++} & TPAMI'21 & ResNet50 & 86.2 & 80.0 & 83.4\\
		    \hline
		    CSL(800) (FPN based) & - & ResNet50 & 84.3 & 83.0 & 83.7\\
		    GCL(800) (FPN based) & - & ResNet50 & 84.7 & 82.6 & 83.6\\
			\bottomrule
	\end{tabular}}
	\label{table:icdar2015}
\end{table}

\begin{table*}[tb!]
	\centering
	\caption{Detection accuracy (AP) on each category of object and overall performance (mAP) on the DOTA-v1.0 test set, using different backbones.}
	\resizebox{1.0\textwidth}{!}{
		\begin{tabular}{r|r|c|c|c|c|c|c|c|c|c|c|c|c|c|c|c|c}
			\toprule
			Method & Backbone &  PL &  BD &  BR &  GTF &  SV &  LV &  SH &  TC &  BC &  ST &  SBF &  RA &  HA &  SP &  HC &  mAP\\
			\hline
			FR-O \cite{xia2018dota} & ResNet101 & 79.09 & 69.12 & 17.17 & 63.49 & 34.20 & 37.16 & 36.20 & 89.19 & 69.60 & 58.96 & 49.4 & 52.52 & 46.69 & 44.80 & 46.30 & 52.93 \\
			IENet \cite{lin2019ienet} & ResNet101 & 80.20 & 64.54 & 39.82 & 32.07 & 49.71 & 65.01 & 52.58 & 81.45 & 44.66 & 78.51 & 46.54 & 56.73 & 64.40 & 64.24 & 36.75 & 57.14 \\
			R-DFPN \cite{yang2018automatic} & ResNet101 & 80.92 & 65.82 & 33.77 & 58.94 & 55.77 & 50.94 & 54.78 & 90.33 & 66.34 & 68.66 & 48.73 & 51.76 & 55.10 & 51.32 & 35.88 & 57.94 \\
			TOSO \cite{feng2020toso} & ResNet101 & 80.17 & 65.59 & 39.82 & 39.95 & 49.71 & 65.01 & 53.58 & 81.45 & 44.66 & 78.51 & 48.85 & 56.73 & 64.40 & 64.24 & 36.75 & 57.92\\
			PIoU \cite{chen2020piou} & DLA-34 \cite{yu2018deep} & 80.9 & 69.7 & 24.1 & 60.2 & 38.3 & 64.4 & 64.8 & 90.9 & 77.2 & 70.4 & 46.5 & 37.1 & 57.1 & 61.9 & 64.0 & 60.5 \\
			R$^2$CNN \cite{jiang2017r2cnn} & ResNet101 & 80.94 & 65.67 & 35.34 & 67.44 & 59.92 & 50.91 & 55.81 & 90.67 & 66.92 & 72.39 & 55.06 & 52.23 & 55.14 & 53.35 & 48.22 & 60.67 \\
			RRPN \cite{ma2018arbitrary} & ResNet101 & 88.52 & 71.20 & 31.66 & 59.30 & 51.85 & 56.19 & 57.25 & 90.81 & 72.84 & 67.38 & 56.69 & 52.84 & 53.08 & 51.94 & 53.58 & 61.01 \\
			Axis Learning \cite{xiao2020axis} & ResNet101 & 79.53 & 77.15 & 38.59 & 61.15 & 67.53 & 70.49 & 76.30 & 89.66 & 79.07 & 83.53 & 47.27 & 61.01 & 56.28 & 66.06 & 36.05 & 65.98\\
			ICN \cite{azimi2018towards} & ResNet101 & 81.40 & 74.30 & 47.70 & 70.30 & 64.90 & 67.80 & 70.00 & 90.80 & 79.10 & 78.20 & 53.60 & 62.90 & 67.00 & 64.20 & 50.20 & 68.20 \\
			RADet \cite{li2020radet} & ResNeXt101 \cite{xie2017aggregated} & 79.45 & 76.99 & 48.05 & 65.83 & 65.46 & 74.40 & 68.86 & 89.70 & 78.14 & 74.97 & 49.92 & 64.63 & 66.14 & 71.58 & 62.16 & 69.09 \\
			RoI-Transformer \cite{ding2018learning} & ResNet101 & 88.64 & 78.52 & 43.44 & 75.92 & 68.81 & 73.68 & 83.59 & 90.74 & 77.27 & 81.46 & 58.39 & 53.54 & 62.83 & 58.93 & 47.67 & 69.56 \\
			P-RSDet \cite{zhou2020arbitrary}	& ResNet101 & 89.02 & 73.65 & 47.33 & 72.03 & 70.58 & 73.71 & 72.76 & 90.82 & 80.12 & 81.32 & 59.45 & 57.87 & 60.79 & 65.21 & 52.59 & 69.82 \\
			CAD-Net \cite{zhang2019cad} & ResNet101 & 87.8 & 82.4 & 49.4 & 73.5 & 71.1 & 63.5 & 76.7 & 90.9 & 79.2 & 73.3 & 48.4 & 60.9 & 62.0 & 67.0 & 62.2 & 69.9 \\
			O$^2$-DNet \cite{wei2020oriented} & Hourglass104 \cite{newell2016stacked} & 89.31 & 82.14 & 47.33 & 61.21 & 71.32 & 74.03 & 78.62 & 90.76 & 82.23 & 81.36 & 60.93 & 60.17 & 58.21 & 66.98 & 61.03 & 71.04 \\
			AOOD \cite{zou2020arbitrary} & DPN \cite{chen2017dual} & 89.99 & 81.25 & 44.50 & 73.20 & 68.90 & 60.33 & 66.86 & 90.89 & 80.99 & 86.23 & 64.98 & 63.88 & 65.24 & 68.36 & 62.13 & 71.18 \\
			Cascade-FF \cite{hou2020cascade} & ResNet152 & 89.9 & 80.4 & 51.7 & \textbf{77.4} & 68.2 & 75.2 & 75.6 & 90.8 & 78.8 & 84.4 & 62.3 & 64.6 & 57.7 & 69.4 & 50.1 & 71.8 \\
			BBAVectors \cite{yi2021oriented} & ResNet101 & 88.35 & 79.96 & 50.69 & 62.18 & 78.43 & 78.98 & \textbf{87.94} & 90.85 & 83.58 & 84.35 & 54.13 & 60.24 & 65.22 & 64.28 & 55.70 & 72.32 \\
			SCRDet \cite{yang2019scrdet} & ResNet101 & 89.98 & 80.65 & 52.09 & 68.36 & 68.36 & 60.32 & 72.41 & 90.85 & \textbf{87.94} & 86.86 & 65.02 & 66.68 & 66.25 & 68.24 & 65.21 & 72.61\\
			SARD \cite{wang2019sard} & ResNet101 & 89.93 & 84.11 & 54.19 & 72.04 & 68.41 & 61.18 & 66.00 & 90.82 & 87.79 & 86.59 & 65.65 & 64.04 & 66.68 & 68.84 & 68.03 & 72.95 \\
			GLS-Net \cite{li2020object} & ResNet101 & 88.65 & 77.40 & 51.20 & 71.03 & 73.30 & 72.16 & 84.68 & 90.87 & 80.43 & 85.38 & 58.33 & 62.27 & 67.58 & 70.69 & 60.42 & 72.96 \\
			DRN \cite{pan2020dynamic} & Hourglass104 & 89.71 & 82.34 & 47.22 & 64.10 & 76.22 & 74.43 & 85.84 & 90.57 & 86.18 & 84.89 & 57.65 & 61.93 & 69.30 & 69.63 & 58.48 & 73.23 \\
			FADet \cite{li2019feature} & ResNet101 & \textbf{90.21} & 79.58 & 45.49 & 76.41 & 73.18 & 68.27 & 79.56 & 90.83 & 83.40 & 84.68 & 53.40 & 65.42 & 74.17 & 69.69 & 64.86 & 73.28\\
			MFIAR-Net \cite{yang2020multi} & ResNet152 & 89.62 & 84.03 & 52.41 & 70.30 & 70.13 & 67.64 & 77.81 & 90.85 & 85.40 & 86.22 & 63.21 & 64.14 & 68.31 & 70.21 & 62.11 & 73.49 \\
			R$^3$Det \cite{yang2021r3det} & ResNet152 & 89.49 & 81.17 & 50.53 & 66.10 & 70.92 & 78.66 & 78.21 & 90.81 & 85.26 & 84.23 & 61.81 & 63.77 & 68.16 & 69.83 & 67.17 & 73.74 \\
			RSDet \cite{qian2021learning} & ResNet152 & 90.1 & 82.0 & 53.8 & 68.5 & 70.2 & 78.7 & 73.6 & \textbf{91.2} & 87.1 & 84.7 & 64.3 & 68.2 & 66.1 & 69.3 & 63.7 & 74.1 \\ 
			Gliding Vertex \cite{xu2020gliding} & ResNet101 & 89.64 & 85.00 & 52.26 & 77.34 & 73.01 & 73.14 & 86.82 & 90.74 & 79.02 & 86.81 & 59.55 & \textbf{70.91} & 72.94 & 70.86 & 57.32 & 75.02 \\
			Mask OBB \cite{wang2019mask} & ResNeXt-101 & 89.56 & \textbf{85.95} & 54.21 & 72.90 & 76.52 & 74.16 & 85.63 & 89.85 & 83.81 & 86.48 & 54.89 & 69.64 & 73.94 & 69.06 & 63.32 & 75.33 \\
			FFA \cite{fu2020rotation} & ResNet101 & 90.1 & 82.7 & 54.2 & 75.2 & 71.0 & 79.9 & 83.5 & 90.7 & 83.9 & 84.6 & 61.2 & 68.0 & 70.7 & 76.0 & 63.7 & 75.7 \\
			APE \cite{zhu2020adaptive} & ResNeXt-101 & 89.96 & 83.62 & 53.42 & 76.03 & 74.01 & 77.16 & 79.45 & 90.83 & 87.15 & 84.51 & \textbf{67.72} & 60.33 & \textbf{74.61} & 71.84 & 65.55 & 75.75 \\
			CenterMap OBB \cite{wang2020learning} & ResNet101 & 89.83 & 84.41 & \textbf{54.60} & 70.25 & 77.66 & 78.32 & 87.19 & 90.66 & 84.89 & 85.27 & 56.46 & 69.23 & 74.13 & 71.56 & 66.06 & 76.03\\
			\hline
			CSL & ResNet152 & 89.33 & 84.88 & 53.70 & 75.68 & 77.57 & 80.21 & 84.18 & 89.80 & 86.57 & 86.22 & 71.86 & 64.48 & 73.48 & \textbf{74.84} & 66.05 & 77.26 \\
			GCL & ResNet152 & 89.26 & 83.59 & 53.05 & 72.76 & 78.13 & 81.97 & 86.94 & 90.36 & 85.98 & 86.94 & 66.19 & 65.56 & 73.29 & 70.56 & \textbf{69.99} & 76.97 \\
		    BCL & ResNet152 & 89.32 & 83.54 & 53.60 & 72.70 & \textbf{78.94} & \textbf{82.66} & 87.27 & 90.69 & 86.61 & \textbf{87.98} & 66.49 & 66.97 & 73.20 & 70.65 & 69.90 & \textbf{77.37}\\
			\bottomrule
	\end{tabular}}
	\label{table:DOTA_OBB}
	\vspace{-8pt}
\end{table*}

\noindent \textbf{Visual analysis of angular features.}
By zooming in on part of Figure \ref{fig:csl-gauss-180}, we show that the prediction of the boundary conditions become continuous (for example, two large vehicle in the same direction predicted $90^\circ$ and $-88^\circ$, respectively). This phenomenon reflects the purpose of designing the CSL: the labels are periodic (circular) and the prediction of adjacent angles has a certain tolerance. In order to confirm that the angle classifier has indeed learned this property, we visually analyze the angular features of each region of interest (RoI) in the FPN detector by principal component analysis (PCA) \cite{wold1987principal}, as shown in Figure \ref{fig:vis_feature}. The detector does not learn the orientation information well when the pulse window function is used. It can be seen from the first row of Figure \ref{fig:vis_feature} that the feature distribution of RoI is relatively random, and the prediction results of some angles occupy the vast majority. For the Gaussian function, the feature distribution is obvious a ring structures, and the features of adjacent angles are close to each other and have a certain overlap. It is this property that helps CSL-based detectors to eliminate boundary problems and accurately obtain the orientation and scale information. We also show the visualization results when the number of angle categories are 4 and 8, as shown in Figure \ref{fig:feature_vis_dcl}. %This proves that it is feasible to use classification for orientation estimation, even if only the simplest cross-entropy loss function is used.

\subsection{Comparison with the State-of-the-Art Methods}\label{sec:sota}
\noindent \textbf{Results on HRSC2016.}
The HRSC2016 contains lots of large aspect ratio ship instances with arbitrary orientation, which poses a huge challenge to the positioning accuracy of the detector. Table \ref{table:HRSC2016} shows that our models achieve superior performances, about 89.62\% (96.10\%), 89.56\% (96.02\%) and 89.46\% (96.41\%), for CSL, GCL and BCL respectively.\\

\noindent \textbf{Results on ICDAR2015.}
Scene text detection has been a well-studied field and many advanced techniques are specifically designated to texts while our proposed model is general for rotation detection. Moreover, to achieve competitive performance in text detection, it often further involves non-trivial processing and like using external data, such as RRPN \cite{ma2018arbitrary}, PAN \cite{wang2019efficient}, TextField \cite{xu2019textfield}, FOTS \cite{liu2018fots} and TextDragon \cite{feng2019textdragon}, powerful pre-trained weights on SynthText \cite{gupta2016synthetic} or COCO \cite{lin2014microsoft} (e.g. SBD \cite{liu2019omnidirectional}, SegLink \cite{shi2017detecting}, TextField \cite{xu2019textfield}, DBNet \cite{liao2020real}, PAN++ \cite{wang2021pan++}, PolarMask++ \cite{xie2021polarmask++}), model ensemble (e.g. Inceptext \cite{yang2018inceptext}). so we are refrained to over compare with state-of-the-art text detection models tailored to texts, and become more focused on the proposed components themselves which we do not want to couple with other factors. Table \ref{table:icdar2015} shows the comparative experiments on the ICDAR2015, and our methods obtain the competitive F-measure without using external data or powerful pre-trained weight.

\noindent \textbf{Results on DOTA-v1.0.}
We choose DOTA-v1.0 as the main validation dataset due to the complexity of the remote sensing and the large number of small, cluttered and rotated objects in the dataset. The used data augmentation include random horizontal, vertical flipping, random graying, and random rotation. Training and testing scale is set to $[400, 600, 720, 800, 1000, 1100]$. As shown in Table \ref{table:DOTA_OBB}, CSL-based, GCL-based, BCL-based protocols show competitive performance, at the accuracy of 77.26\%, 76.97\% and 77.37\%, respectively.

\begin{table}[tb!]
	\centering
	\caption{Detection accuracy on each object (AP) and overall performance (mAP) on OHD-SJTU-S. `H' and `R' represent the horizontal and rotating anchors, respectively. Here the numbers in the subscript of AP i.e. 50, 75, 95 represent the threshold of IoU. Note our model's performance gain is even pronounced on the challenging AP$_{75}$ and AP$_{50:95}$ metrics.}
	\resizebox{0.48\textwidth}{!}{
		\begin{tabular}{l|cc|ccc}
			\toprule
			
			Method & PL (AP$_{50}$) & SH (AP$_{50}$) & AP$_{50}$ & AP$_{75}$ & AP$_{50:95}$ \\
			
			\hline
			R$^2$CNN \cite{jiang2017r2cnn} & \textbf{90.91} & 77.66 & 84.28 & 55.00 & 52.80 \\
			RRPN \cite{ma2018arbitrary} & 90.14 & 76.13 & 83.13 & 27.87 & 40.74 \\
			RetinaNet-H \cite{yang2021r3det} & 90.86 & 66.32 & 78.59 & 58.45 & 53.07 \\
			RetinaNet-R \cite{yang2021r3det} & 90.82 & \textbf{88.14} & \textbf{89.48} & 74.62 & 61.86 \\
			% RoI-Transformer \cite{ding2018learning} & 90.91 & 88.32 & 89.61 & 86.69 & 67.57 \\
			R$^3$Det \cite{yang2021r3det} & 90.82 & 85.59 & 88.21 & 67.13 & 56.19 \\
			\hline
		    OHDet (ours) & 90.74 & 87.59 & 89.06 & \textbf{78.55} & \textbf{63.94}\\
			\bottomrule
	\end{tabular}}
	\label{table:OHD-SJTU-S}
\end{table}

\begin{table}[tb!]
	\centering
	\caption{Detection accuracy on each object (AP) and overall performance (mAP) on OHD-SJTU-L. Note our model's performance gain is even pronounced on the challenging AP$_{75}$ and AP$_{50:95}$ metrics.}
	\resizebox{0.48\textwidth}{!}{
		\begin{tabular}{l|cccccc|ccc}
			\toprule
			
			Method & PL & SH & SV & LV & HA & HC & AP$_{50}$ & AP$_{75}$ & AP$_{50:95}$ \\
			
			\midrule
			R$^2$CNN \cite{jiang2017r2cnn} & 90.02 & 80.83 & 63.07 & 64.16 & \textbf{66.36} & 55.94 & 70.06 & 32.70 & 35.44 \\
			RRPN \cite{ma2018arbitrary} & 89.55 & 82.60 & 57.36 & 72.26 & 63.01 & 45.27 & 68.34 & 22.03 & 31.12 \\
			RetinaNet-H \cite{yang2021r3det} & \textbf{90.22} & 80.04 & 63.32 & 63.49 & 63.73 & 53.77 & 69.10 & 35.90 & 36.89 \\
			RetinaNet-R \cite{yang2021r3det} & 90.00 & 86.90 & 63.24 & \textbf{86.90} & 62.85 & 52.35 & \textbf{72.78} & 40.13 & 40.58 \\
			R$^3$Det \cite{yang2021r3det} & 89.89 & \textbf{87.69} & \textbf{65.20} & 78.95 & 57.06 & 53.50 & 72.05 & 36.51 & 38.57 \\
			\hline
		    OHDet (ours) & 89.73 & 86.63 & 61.37 & 78.80 & 63.76 & \textbf{54.62} & 72.49 & \textbf{43.60} & \textbf{41.29}\\
			\bottomrule
	\end{tabular}}
	\label{table:OHD-SJTU-L}
\end{table}

\begin{figure*}[tb!]
	\centering
	\subfigure[Ship]{
		\begin{minipage}[t]{0.58\linewidth}
			\centering
			\includegraphics[width=0.97\linewidth]{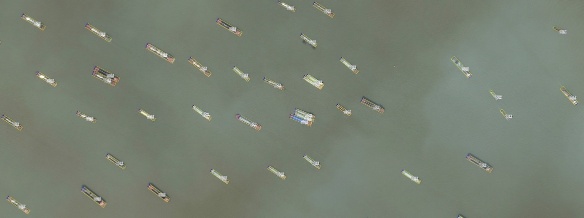}
			%\caption{fig1}
		\end{minipage}%
		\label{fig:ship_head_6}
	}
	\subfigure[Harbor]{
		\begin{minipage}[t]{0.32\linewidth}
			\centering
			\includegraphics[width=0.97\linewidth]{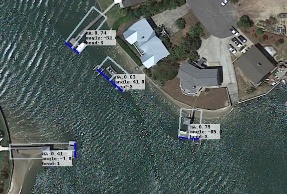}
			%\caption{fig2}
		\end{minipage}
		\label{fig:harbor_head_3}
	} \\
	\subfigure[Vehicle]{
		\begin{minipage}[t]{0.45\linewidth}
			\centering
			\includegraphics[width=0.97\linewidth]{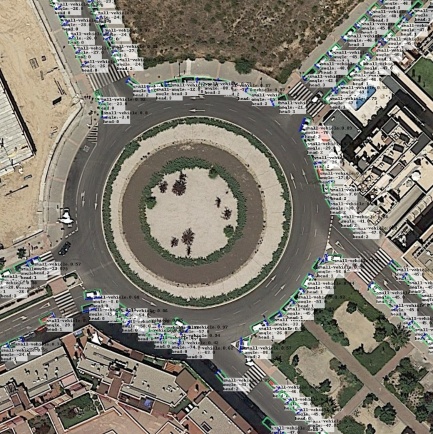}
			%\caption{fig2}
		\end{minipage}
		\label{fig:car_head_4}
	}
	\subfigure[Plane]{
		\begin{minipage}[t]{0.45\linewidth}
			\centering
			\includegraphics[width=0.97\linewidth]{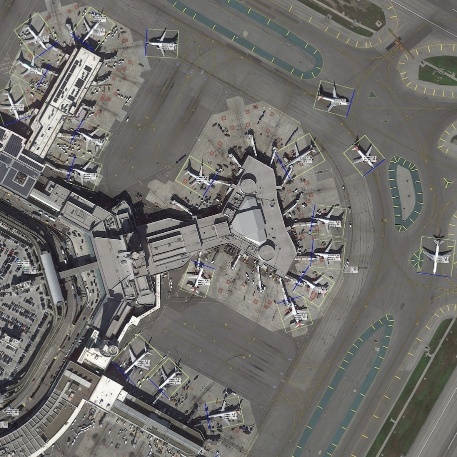}
			%\caption{fig2}
		\end{minipage}
		\label{fig:plane_head_5}
	}
	\centering
	\vspace{-8pt}
	\caption{Detection examples of our  method in large-scale scenarios on OHD-SJTU. Our method can both effectively handle the dense and rotating cases. The blue border in the bounding box denotes the predicted head of the object.}
	\label{fig:head_vis}
\end{figure*}

\noindent \textbf{Results on OHD-SJTU.}
We assess the performance of state-of-the-art rotation object detection methods on OHD-SJTU, mainly include R$^2$CNN, RRPN, RetinaNet, R$^3$Det.
All experiments are based on the same setting, using ResNet101 as the backbone. Except for data augmentation (include random horizontal, vertical flipping, random graying, and random rotation) is used in OHD-SJTU-S, no other tricks are used. As shown in Table \ref{table:OHD-SJTU-S}, the large number of dense ship with large aspect ratio in the OHD-SJTU-S brings huge challenges to the high-precision detection capabilities of the detector. RetinaNet-R and R$^3$Det use rotating anchor and cascade structure respectively, which makes them stand out in high-precision indicators, such as AP$_{75}$. OHDet uses CSL and IoU smooth L1 in combination, which not only reduces the amount of model parameters, but also avoids the side effects of boundary problems. Then OHDet is further combined with the cascade structure to achieve the best detection performance, at about 63.94\%.
Especially in AP$_{75}$ and AP$_{50:95}$, our method is 3.93\% and 2.08\% higher than the second-best method. Similar conclusions can also be obtained from OHD-SJTU-L. As shown in Table \ref{table:OHD-SJTU-L}, OHDet achieves a notable advantage in high-precision indicators and achieves the best performance (about 41.29\%).

\subsection{Object Heading Detection Experiment}\label{sec:ohdet_exp}
For rotation object detection, the expected output includes the object category and the IoU between the predicted bounding box and ground truth, which is often compared with a certain threshold, e.g. 0.5. In contrast, object heading detection additionally outputs the head prediction results. Three indicators are used to measure the performance: OBB mAP, OHD mAP and Head Accuracy. OBB mAP is a detection metric without considering object head prediction, which is consistent with rotation detection. While OHD mAP additionally considers the accuracy of object head prediction. Therefore, for the same model, the upper bound of OHD mAP is OBB mAP when the head prediction accuracy is 100\%. The Head Accuracy indicates the accuracy of head prediction in all the detection boxes judged as true positive (TP). Figure \ref{fig:head_vis} visualizes the object heading detection on different categories.

\begin{figure}[tb!]
	\centering
	\subfigure[Habor and Helicopter]{
		\begin{minipage}[t]{0.21\linewidth}
			\centering
			\includegraphics[width=0.97\linewidth]{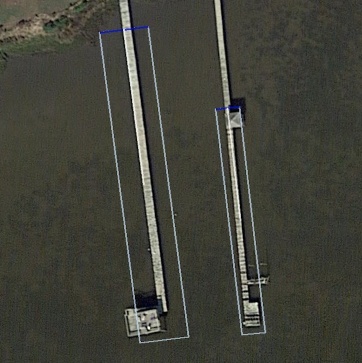}
			%\caption{fig1}
		\end{minipage}%
		\label{fig:HA_bad_case}
		\begin{minipage}[t]{0.21\linewidth}
			\centering
			\includegraphics[width=0.97\linewidth]{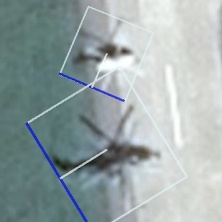}
			%\caption{fig2}
		\end{minipage}
		\label{fig:HC_bad_case}
	}
	\subfigure[Large Vehicle and Plane]{
		\begin{minipage}[t]{0.21\linewidth}
			\centering
			\includegraphics[width=0.97\linewidth]{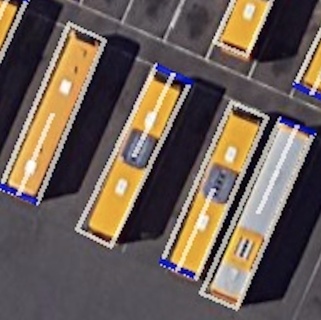}
			%\caption{fig1}
		\end{minipage}%
		\label{fig:LV_bad_case}
		\begin{minipage}[t]{0.21\linewidth}
			\centering
			\includegraphics[width=0.97\linewidth]{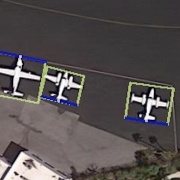}
			%\caption{fig2}
		\end{minipage}
		\label{fig:PL_bad_case}
	}\\
	\vspace{-5pt}
	\subfigure[Ship]{
		\begin{minipage}[t]{0.21\linewidth}
			\centering
			\includegraphics[width=0.97\linewidth]{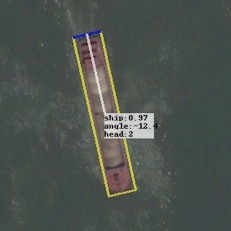}
			%\caption{fig2}
		\end{minipage}
		\label{fig:SH_bad_case}
		\begin{minipage}[t]{0.21\linewidth}
			\centering
			\includegraphics[width=0.97\linewidth]{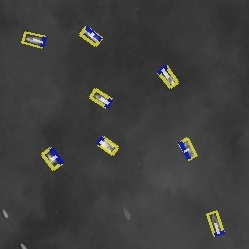}
			%\caption{fig2}
		\end{minipage}
		\label{fig:SH_bad_case_1}
	}
	\subfigure[Small Vehicle]{
		\begin{minipage}[t]{0.21\linewidth}
			\centering
			\includegraphics[width=0.97\linewidth]{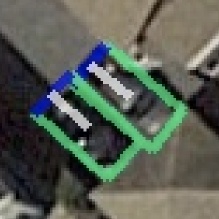}
			%\caption{fig2}
		\end{minipage}
		\label{fig:SV_bad_case}
		\begin{minipage}[t]{0.21\linewidth}
			\centering
			\includegraphics[width=0.97\linewidth]{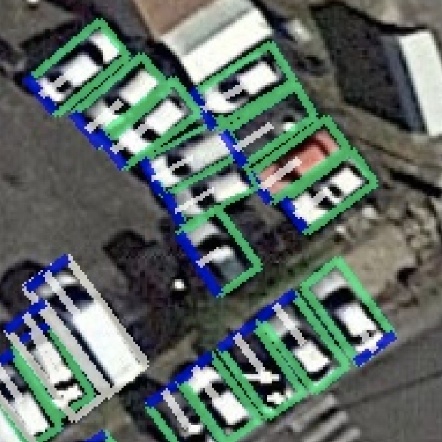}
			%\caption{fig2}
		\end{minipage}
		\label{fig:SV_bad_case_1}
	}
	\centering
	\caption{Illustration for the main failure cases of head prediction: extremely similar head and tail, fuzzy small object, densely arranged, and small sample size, etc.}
	\label{fig:bad_case}
\end{figure}

Table \ref{table:OHD-eval} shows the performance of OHDet on the two sub-data sets of OHD-SJTU. Due to the high image resolution and clear objects in OHD-SJTU-S, a very high head prediction accuracy can be achieved: 94.25\%. However, head prediction still faces challenges in complex environments. For example, the head prediction accuracy is only 64.12\% on OHD-SJTU-L dataset, and the detection performance drops from 37.66\% to 22.97\%. It can be seen that the addition of head prediction conditions will greatly reduce the detection performance. The main difficulties of inaccurate head prediction are as follows: extremely similar head and tail (e.g. LV, SH, SV), fuzzy small object (e.g. SV, SH), densely arranged (e.g. LV, SH, SV), and small sample size (e.g. HC). Figure \ref{fig:bad_case} shows some bad cases. We hope that the open source of the OHD-SJTU can promote the research of related methods.

\section{Conclusion}\label{sec:conclusion}
In this paper, we have particularly identified the boundary problems as faced by different regression-based rotation detection methods. The main cause of boundary problems based on regression methods is that the ideal predictions are beyond the defined range. Therefore, considering the prediction of the object angle as a classification problem to better limit the prediction results, and then we design a Circular Smooth Label (CSL) to adapt to the periodicity of the angle and increase the tolerance of classification between adjacent angles with little accuracy error. We also introduce four window functions in CSL and explore the effect of different window radius sizes on detection performance. 

\begin{table}[tb!]
\centering
\caption{Results of object heading detection on OHD-SJTU dataset, covering six categories under three different settings of IoU.}
\resizebox{0.48\textwidth}{!}{
	\begin{tabular}{cc|cccccc|ccc}
		\toprule
		
		Tag & Task & PL & SH & SV & LV & HA & HC & IoU$_{50}$ & IoU$_{75}$ & IoU$_{50:95}$ \\
		
		\hline
		\multirow{3}{*}{S}
	    & OBB & 90.73 & 88.59 & -- & -- & -- & -- & 89.66 & 75.62 & 61.49\\
	    & OHD & 76.89 & 86.40 & -- & -- & -- & -- & 81.65 & 65.51 & 55.09\\
	    & Head & 90.91 & 94.87 & -- & -- & -- & -- & 92.89 & 93.81 & 94.25\\
	    \hline
	    \multirow{3}{*}{L}
	    & OBB & 89.62 & 85.58 & 48.45 & 76.55 & 61.43 & 33.87 & 65.92 & 38.80 & 37.66\\
	    & OHD & 59.93 & 47.57 & 26.59 & 35.32 & 41.29 & 17.53 & 38.04 & 24.86 & 22.97\\
	    & Head & 74.43 & 68.39 & 60.15 & 57.79 & 76.66 & 49.06 & 64.41 & 65.17 & 64.12\\
		\bottomrule
\end{tabular}}
\label{table:OHD-eval}
\end{table}

To reduce the excessive model parameters caused by CSL, we further design a Densely Coded Label (DCL), which greatly reduces the length of the encoding while ensuring that the angle prediction accuracy is not reduced. An angle fine-tuning mechanism is also devised to eliminate the theoretical prediction errors caused by angle dispersion which has been a common issue in whatever CSL and DCL. Our devised angle high-precision classification is also the first application in rotation detection. 

We further fulfill the function of object heading detection, called OHDet, which is used to find the head of object. We annotate and release a dataset for rotation detection and object heading detection, called OHD-SJTU. Extensive experiments and visual analysis on different detectors and datasets show the effectiveness of our approach.

	%\appendices
	%% you can choose not to have a title for an appendix
	%% if you want by leaving the argument blank
	%\section{}

	\section*{Acknowledgment}
    This work was partly supported by National Key Research and Development Program of China (2020AAA0107600), Shanghai Municipal Science and Technology Major Project (2021SHZDZX0102) and National Natural Science Foundation of China (U20B2068, 61972250). Xue Yang is partly supported by Wu Wen Jun Honorary Doctoral Scholarship, AI Institute, Shanghai Jiao Tong University.
	
	\ifCLASSOPTIONcaptionsoff
	\newpage
	\fi

	\bibliographystyle{IEEEtran}
	\bibliography{egbib}
	
\end{document}